\documentclass{article} 
\usepackage[final]{colm2025_conference}

\usepackage{microtype}
\usepackage{hyperref}
\usepackage{url}
\usepackage{booktabs}
\usepackage{graphicx}
\usepackage{subfig}
\usepackage{caption}
\usepackage{tikz}
\usetikzlibrary{trees, positioning} 
\usepackage{algorithm}
\usepackage{algpseudocode}
\usepackage{longtable}
\usepackage{pifont}
\usepackage{arydshln}
\usepackage{multirow}
\usepackage{wrapfig}

\usepackage{lineno}

\definecolor{darkblue}{rgb}{0, 0, 0.5}
\hypersetup{colorlinks=true, citecolor=darkblue, linkcolor=darkblue, urlcolor=darkblue}

\title{FineMedLM-o1: Enhancing Medical Knowledge Reasoning Ability of LLM from Supervised Fine-Tuning to Test-Time Training}

\author{Hongzhou Yu$^{1}$, Tianhao Cheng$^1$, Yingwen Wang$^2$, Wen He$^2$, Qing Wang$^2$,\\ \textbf{Ying Cheng$^{1}\thanks{\ \ Corresponding Authors}$     , Yuejie Zhang$^1$, Rui Feng$^{1\dagger}$, Xiaobo Zhang$^{2\dagger}$} \\
  $^1$Fudan University \ \  $^2$Children’s Hospital of Fudan University\\
  \texttt{\{hzyu24,thcheng23\}@m.fudan.edu.cn}, \\
  \texttt{\{yingwenwang,hewen,chengy18,yjzhang,fengrui\}@fudan.edu.cn},\\\,
  \texttt{zhangxiaobo0307@163.com}}

\begin{document}

\ifcolmsubmission
\linenumbers
\fi

\maketitle

\begin{abstract}
Recent advancements in large language models (LLMs) have shown promise in medical applications such as disease diagnosis and treatment planning. However, most existing medical LLMs struggle with the deep reasoning required for complex medical problems, such as differential diagnosis and medication recommendations. We propose FineMedLM-o1, which leverages high-quality medical synthetic data and long-form reasoning data for Supervised Fine-Tuning (SFT) and Direct Preference Optimization (DPO), enabling advanced dialogue and deep reasoning capabilities. Additionally, we introduce Test-Time Training (TTT) in the medical domain for the first time, facilitating domain adaptation and ensuring reliable, accurate reasoning.
Experimental results demonstrate that FineMedLM-o1 achieves a 23\% average performance improvement over prior models on key medical benchmarks. Furthermore, the introduction of TTT provides an additional 14\% performance boost, highlighting its effectiveness in enhancing medical reasoning capabilities.
To support this process, we also propose a novel method for synthesizing medical dialogue. Compared to other open-source datasets, our dataset stands out as superior in both quality and complexity.
The project and data will be released on GitHub\footnote{\url{https://github.com/hongzhouyu/FineMed}}.
\end{abstract}

\section{Introduction}

\begin{figure}[t]
\centering
  \includegraphics[width=\textwidth]{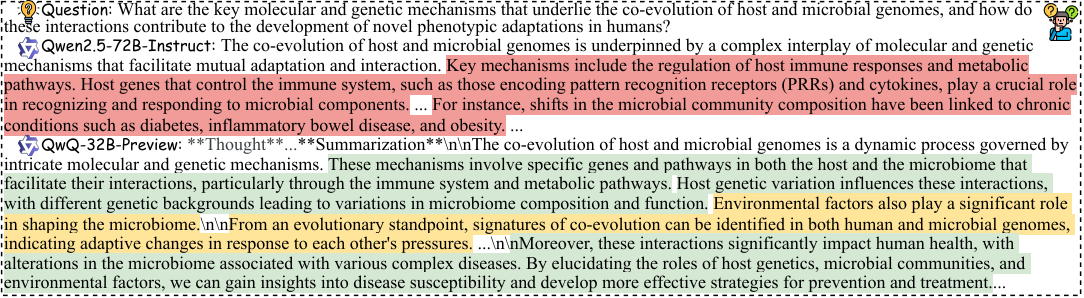}
  \caption{An example from our dataset illustrating the comprehensive reasoning required to provide a reliable answer. The red and green highlights indicate points where both models addressed the same aspects. However, QwQ's response is clearer and more specific. The yellow highlights represent additional content uniquely provided in QwQ's answer.}
  \label{fig:example}
\end{figure}

Medical services are essential benefits that should be accessible to all individuals worldwide, as they contribute to social development and enhance people's overall satisfaction \citep{tian2024chimedgptchinesemedicallarge, zhang2024retfound, ji-etal-2025-robguard}. Recently, LLMs have made significant strides, with several closed-source general LLMs \citep{achiam2023gpt, deepseekai2024deepseekv3technicalreport} achieving impressive performance in medical applications. This progress has inspired the research community to explore the development of more advanced open-source medical LLMs.
Despite these advancements, current medical LLMs still struggle to solve complex problems through deep reasoning. This limitation stems partly from suboptimal training strategies, but more critically, from the inadequacies of existing medical datasets \citep{allenzhu2024physicslanguagemodels31, allenzhu2024physicslanguagemodels33}. These datasets often lack robust logical structures and fail to include essential components such as chain-of-thought (CoT) data or o1-style long-form reasoning data, both of which are crucial for teaching models how to think critically and reason effectively \citep{allenzhu2024physicslanguagemodels32}.
In addition, many complex medical problems cannot be resolved with direct answers alone. Instead, they require comprehensive reasoning to arrive at reliable conclusions. Figure~\ref{fig:example} presents a comparison of the two approaches to solving complex medical problems. As illustrated, responses generated through reasoning are typically clearer, more specific, and better suited to fully addressing the given problem. Without the capacity for thoughtful reasoning, current LLMs are prone to generating incorrect responses, increasing the risk of severe medical errors and potentially leading to critical consequences \citep{tian2024medical}.

Recently, significant efforts have been devoted to enhancing the ability of LLMs in the medical domain to generate reliable and accurate responses. These approaches typically involve using powerful LLMs to rewrite datasets, thereby filling in gaps in the logical content of the corpus, or incorporating prompts to guide the model in step-by-step reasoning \citep{abdin2024phi, xu2023wizardlm, wu2024chainofthoughcotpromptingstrategies}. However, these methods fail to address the core issue that medical LLMs are unable to perform deep reasoning. This limitation persists because existing strategies neither validate the quality and complexity of synthetic data nor integrate o1-style long-form reasoning data into the training process. 
As a result, the generated responses may still lack the depth and rigor required for high-stakes medical decision-making. Addressing this challenge requires a more systematic approach that not only refines data quality but also fosters advanced reasoning capabilities within LLMs.

In this paper, we propose FineMedLM-o1 and a novel synthetic data generation method aimed at enhancing both the reasoning capabilities and domain adaptation of LLMs in medical contexts. FineMedLM-o1 is trained through SFT and reinforcement learning on the base model, while incorporating TTT during inference to further enhance its reasoning capabilities. The synthetic data generation pipeline is a comprehensive system encompassing instruction generation, scoring, filtering, and response generation, ensuring the creation of high-quality, domain-specific datasets that effectively support model training.

Specifically, the training process of FineMedLM-o1 consists of two steps. First, the model undergoes a fine-grained 3-stage SFT using synthetic medical dialogues. To support this training, we develop a high-quality medical SFT dataset called FineMed, generated using the synthetic data method described above. FineMed is of superior quality, having undergone quality scoring and filtering by the LLM-as-a-judge framework during its development. It consists of about 300,000 samples, which are divided into fine-grained subsets to facilitate the 3-stage SFT process. 
Subsequently, the model's reasoning capabilities are enhanced through DPO \citep{rafailov2024direct}. This step involves further fine-tuning with medical data containing complex instructions and o1-style responses, as well as preference learning using common and o1-style responses. In this stage, we utilize 33,000 high-quality DPO pairs to refine the model’s performance.
To enhance reasoning capabilities, we introduce TTT, which enables the model to retrieve and learn from similar data before generating responses. This technique allows the model to better adapt to domain-specific knowledge and reasoning processes, thereby improving the robustness and reliability of its outputs. 
Our synthetic data generation method incorporates a robust verification framework to evaluate data quality, complexity, medical relevance, and specificity. To the best of our knowledge, we are the first to apply o1-style data and TTT in the medical domain to enhance reasoning capabilities and the first to introduce a validation method for synthetic medical data.


The main contributions are as follows:
(1) We introduce a novel framework for generating large-scale, high-quality synthetic SFT data, the first of its kind for medical data, ensuring strict adherence to content, context, quality, and complexity standards.
(2) We implement a complete process, from SFT and DPO to TTT, for the medical LLM FineMedLM-o1, advancing the exploration of LLM reasoning capabilities on complex medical tasks.
(3) We will open-source all the code, datasets, and resources used in this research, with the goal of supporting further research and fostering innovation within the open-source community.

\section{FineMed}


\subsection{Data Synthesis}

\begin{figure}[t]
\centering
   \subfloat[The pipeline of instruction generation]
   {
     \label{fig:first_image}
     \includegraphics[width=\textwidth]{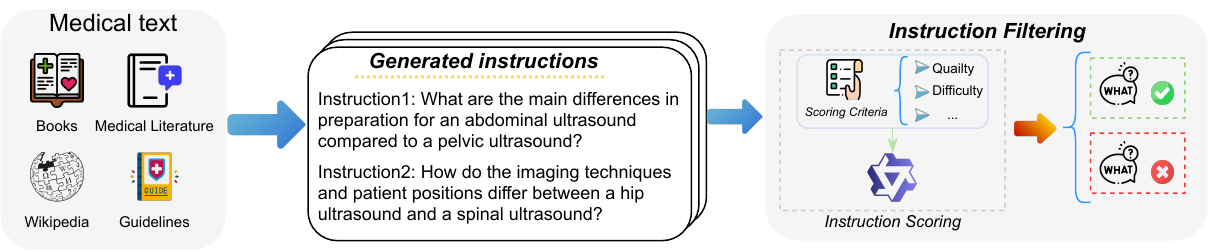}
   }\\[1ex]
   \subfloat[The pipeline of response generation]
   {
     \label{fig:second_image}
     \includegraphics[width=\textwidth]{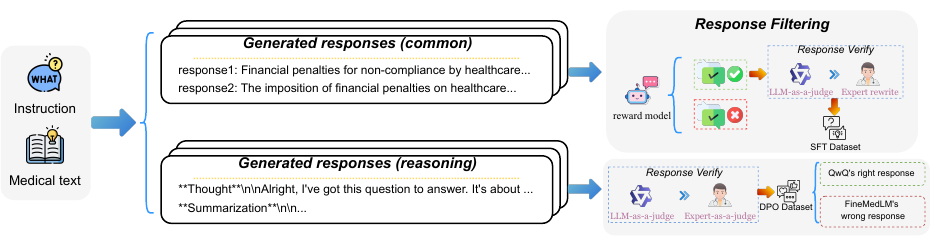}
   }
   \caption{The overall pipeline of generating synthetic data. Figure~\ref{fig:first_image} illustrates the generation of instructions derived from medical texts. Figure~\ref{fig:second_image} depicts the subsequent process of generating common responses and long-form reasoning responses.}
   \label{fig:syntheticdata}
\end{figure}

To construct FineMed, the first step is to generate large amounts of high-quality synthetic medical data. 
In the following, we describe our method, as illustrated in Figure~\ref{fig:syntheticdata}. 

\paragraph{Raw Medical Texts}
Unlike approaches that rely on conversation data from real-world application scenarios on online treatment platforms or filter open-source medical SFT datasets \citep{yang2024pediatricsgpt, gururajan2024aloe}, we aim to use internet corpora (e.g., Common Crawl, CC) as the foundation for our medical knowledge texts. CC inherently includes large-scale question-answer pairs and knowledge-rich textbooks \citep{shao2024deepseekmathpushinglimitsmathematical, yue2024mammoth2}. 
Moreover, advancements in the classification of internet corpora by discipline, such as the open-sourced FineFineWeb \citep{finefineweb}, have significantly accelerated the development of LLMs across various domains. Leveraging this resource, we randomly selected 420,000 samples from the medical subset of FineFineWeb as our raw medical texts.

\paragraph{Instruction Generation}
Building on the perspective proposed by previous researchers \citep{lu2023instag}, we contend that the complexity of a SFT dataset is primarily determined by the intricacy of its instructions. Concurrently, several studies \citep{zhou2024lima, cao2023instruction} have demonstrated that applying quality filtering to synthetic data generated by strong base models can significantly enhance performance by removing low-quality instructions. Based on these insights and with guidance from professional doctors, we establish a set of scoring criteria for generated instructions and utilize Qwen \citep{yang2024qwen2} to evaluate them accordingly. It is noteworthy that we utilize vLLM \citep{kwon2023efficientmemorymanagementlarge} to accelerate the inference process during the generation of synthetic data. 
Specifically, as shown in Figure~\ref{fig:first_image}, we employ Qwen to generate two distinct instructions for each medical text and assign a quality and complexity score to each instruction on a scale of 1 to 10, based on predefined criteria (detailed in Appendix~\ref{sec:a1} and 
 ~\ref{sec:a2}). To ensure the instructions remain relevant to medicine and do not excessively dilute the total quality and complexity scores, relevance to medicine is scored on a scale of 1 to 6. Furthermore, we use Qwen to evaluate whether the generated instructions include specific details from the medical text to prevent unanswerable instructions due to missing contextual information. 
Finally, we filter the scored instructions through a multi-step process, as outlined in Algorithm~\ref{alg:compare_instruction_score} (see Appendix~\ref{sec:appendixb}). This approach yields 333,000 high-quality, high-complexity instruction samples.
 
\paragraph{Response Generation}
As illustrated in Figure~\ref{fig:example}, directly answering high-complexity instructions often leads to fragile responses. To ensure the quality of responses in the SFT dataset, we first categorize instructions into common and complex types based on their complexity scores, with a threshold set at 8. The response generation process is shown in Figure~\ref{fig:second_image}. For common instructions, we use Qwen to generate two stylistically distinct responses and employ a reward model\footnote{\url{https://huggingface.co/sfairXC/FsfairX-LLaMA3-RM-v0.1}\citep{xiong2024iterative} } to select the response that better aligns with human preferences. We then validate the selected responses through a multi-stage process. First, we prompt a LLM to assess whether each response (1) appropriately addresses the given instruction and (2) is grounded in the original source text. Responses that fail either criterion are subsequently reviewed and revised by professional clinicians. Through this pipeline, we curate a high-quality SFT dataset comprising 300,000 instruction–response pairs. For complex instructions, we engage QwQ \citep{team2024qwq} to generate and verify detailed long-form reasoning responses, creating an SFT dataset enriched with o1-style data. Additionally, we use FineMedLM (our SFT-trained model) to generate responses to the same instructions and pair them with QwQ's reasoning outputs, forming a dataset for DPO. Detailed prompts for response generation can be found in Appendix~\ref{sec:a3}.

\subsection{Data classification}

In fine-tuning the base model for medical dialogue applications, we adopt a 3-stage SFT strategy, beginning with a broad medical domain and progressively narrowing the focus to more specific subfields. To achieve this, a fine-grained classification of medical data is essential. We introduce a medical knowledge classification diagram (see Appendix~\ref{sec:appendixc}), which integrates the department structure of Zhongshan Hospital, affiliated with Fudan University\footnote{\url{https://www.zs-hospital.sh.cn/}}, alongside relevant data from the Chinese Hospital Association\footnote{\url{https://www.cha.org.cn/}} (CHA), with additional consultation from professional doctors. Previous studies have shown that providing appropriate prompts enables LLMs to effectively classify data \citep{dong2024survey}. Therefore, we employ Qwen to categorize the medical data according to the aforementioned classification framework, ultimately producing FineMed. The classification prompt used is detailed in Appendix~\ref{sec:appendixd}. FineMed is a SFT dataset comprising five primary and twenty-nine secondary categories. Key statistics of the dataset are summarized in Table~\ref{tab:statistics} (see Appendix~\ref{sec:appendixe} for details). In Section \ref{sec:DatasetStatistics}, we compare the quality and complexity of FineMed with other widely used open-source medical SFT datasets and demonstrate the distribution of data across departments in the semantic space, highlighting the effectiveness of our prompts used for the classification approach.

\subsection{Dataset Analysis} \label{sec:DatasetStatistics}

\begin{figure}[t]
   \centering
   \subfloat[]
   {
     \label{fig:symbolic_reasoning_num_examples}\includegraphics[width=0.48\textwidth]{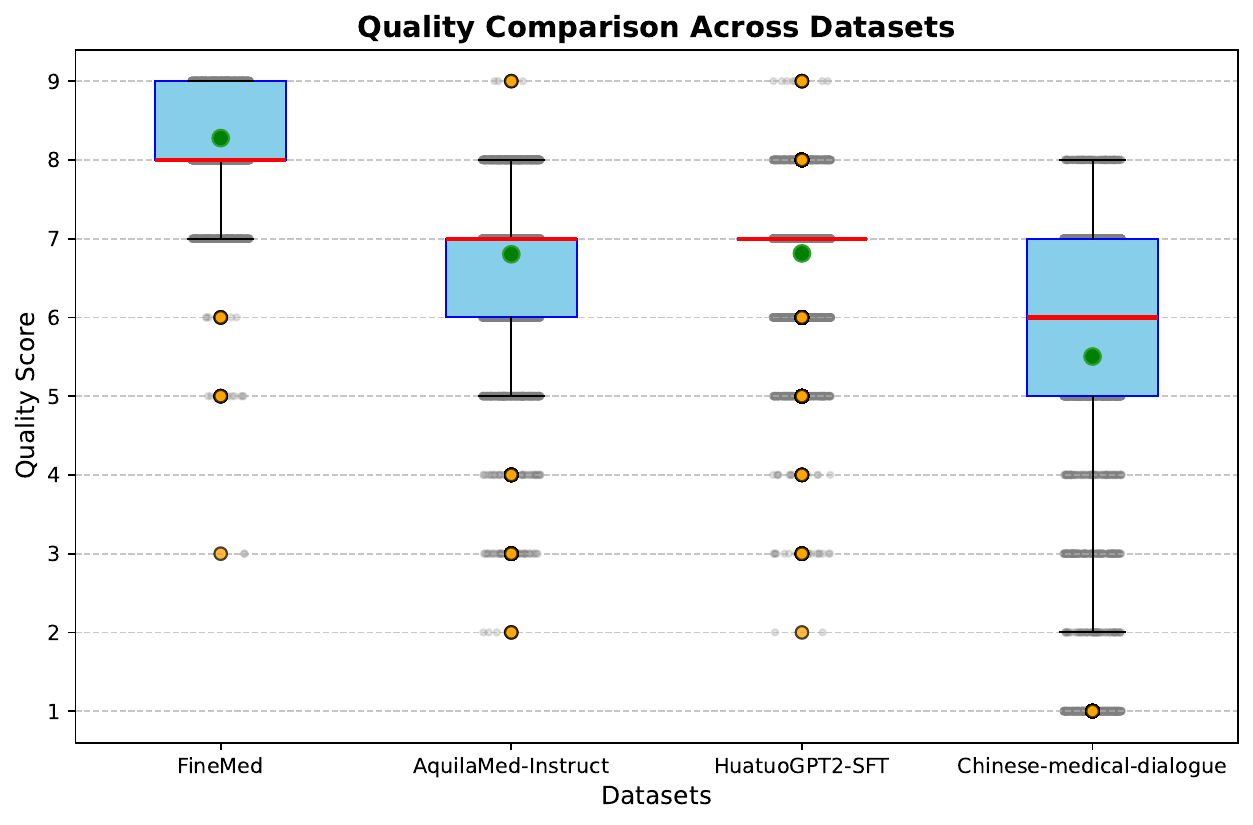}
   }
   \subfloat[]
   {
     \label{fig:arithmetic_reasoning_num_examples}\includegraphics[width=0.48\textwidth]{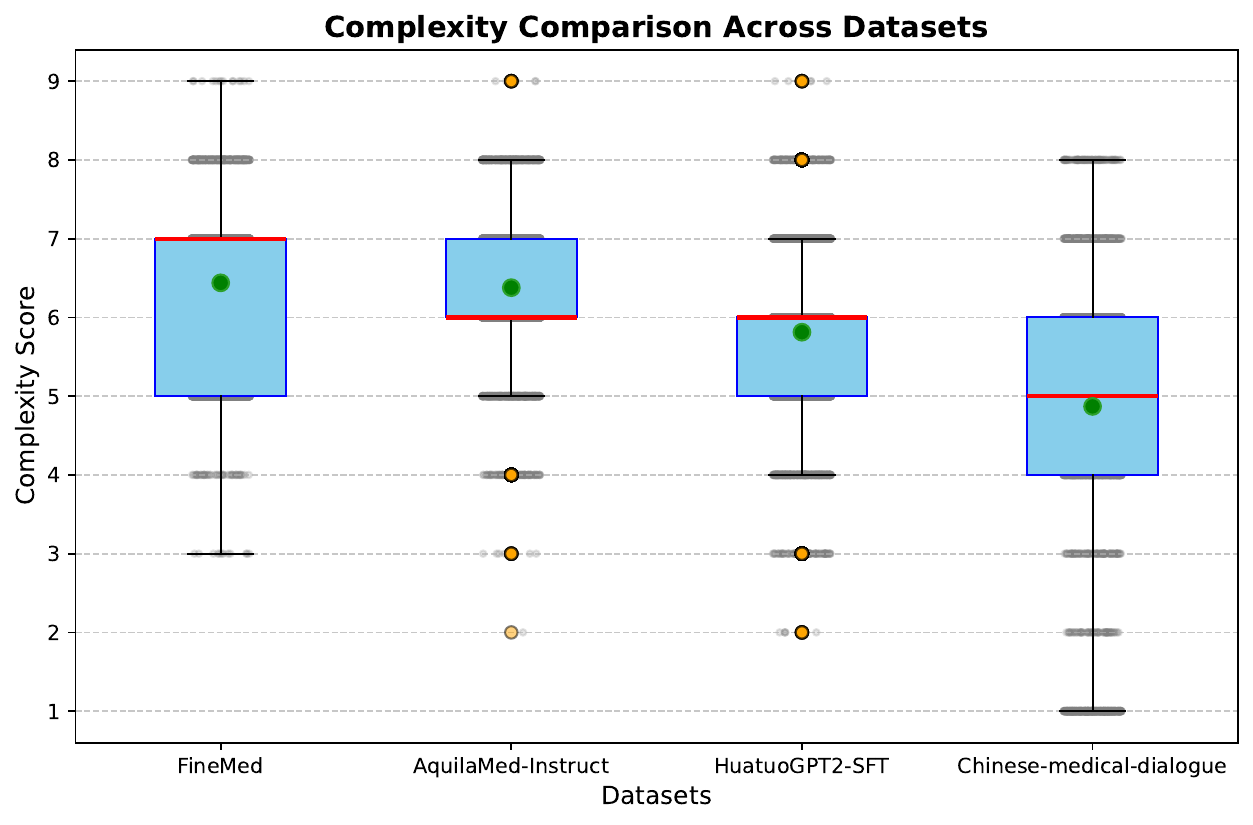}
   }
   \caption{Comparison of four datasets in terms of quality and complexity.}
   \label{fig:compare}
 \end{figure}

\paragraph{Comparison with other datasets}
To better demonstrate the advantages of our proposed synthetic data approach, we employ the LLM-as-a-judge approach to evaluate the quality and complexity of both FineMed and other medical SFT datasets. 
As illustrated in Figure~\ref{fig:compare}, we randomly select 5,000 samples each from FineMed, AquilaMed-Instruct \citep{zhao2024aqulia}, HuatuoGPT2-SFT \citep{chen2023huatuogpt} and Chinese-med-dialogue\footnote{\url{https://huggingface.co/datasets/ticoAg/Chinese-medical-dialogue}} for comparison. In terms of instruction quality, FineMed achieves the highest average quality and median scores, with a relatively concentrated quality distribution. Regarding instruction complexity, both FineMed and AquilaMed-Instruct, which employs the Deita \citep{liu2023makes} method for complexity filtering, demonstrate higher levels of complexity. This result indicates that our scoring criteria effectively capture the impact of the Deita method, with both datasets exhibiting comparable performance in this aspect.

\begin{wrapfigure}{r}{0.45\textwidth}
    \centering
    \includegraphics[width=0.45\textwidth]{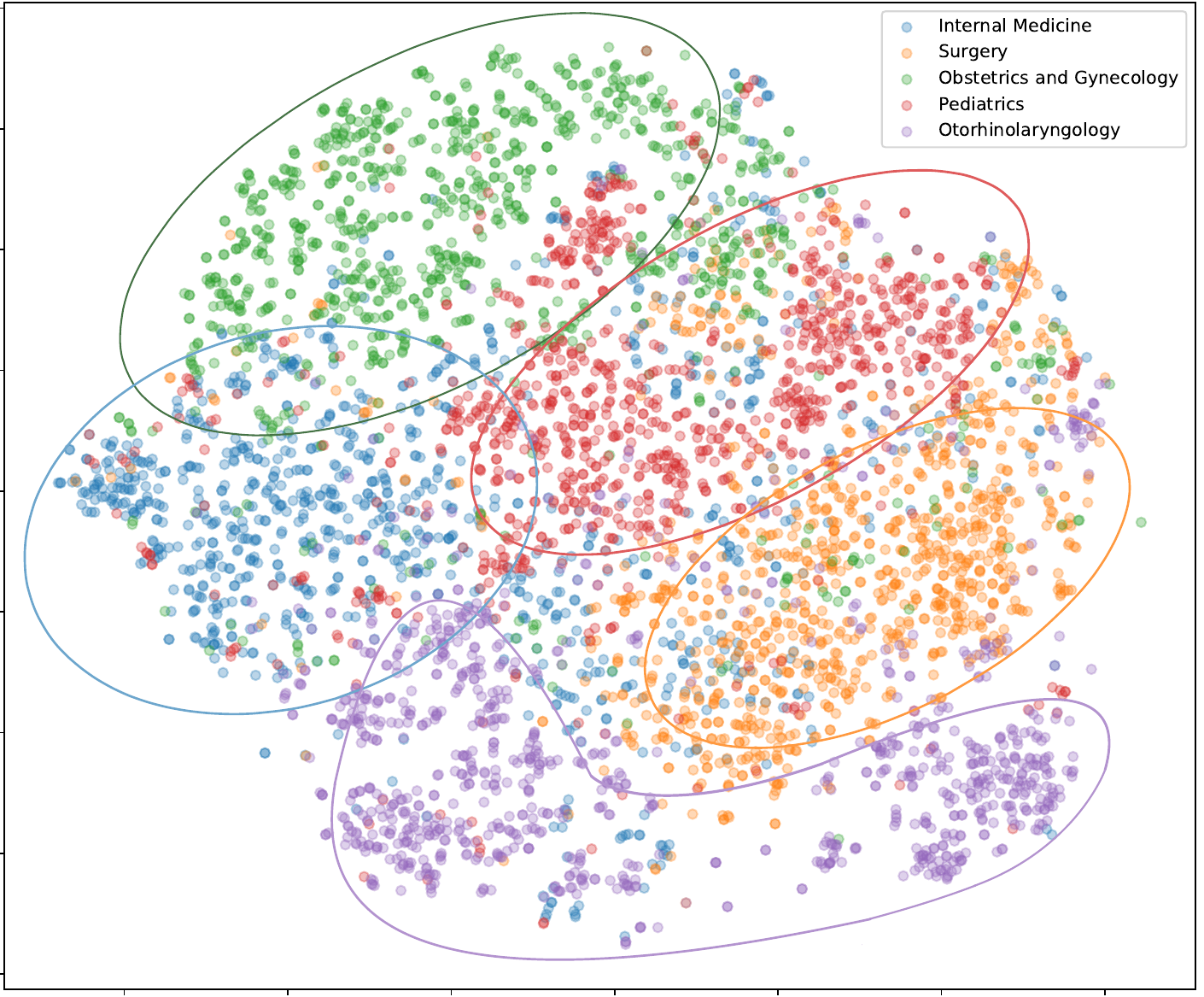}
    \caption{Visualization on the t-SNE data distributions of FineMed.}
    \label{fig:distribution}
\end{wrapfigure}

\paragraph{Verifying the robustness of classification approach}
To facilitate the analysis of data distribution across different first-level departments within FineMed, we employ a two-dimensional semantic space for visualization. This approach enables a more intuitive examination of the relationships among departmental data points, thereby providing empirical support for the validity of our classification methodology and the proposed diagram. Specifically, we leverage the t-distributed Stochastic Neighbor Embedding (t-SNE) technique to project the embeddings of FineMed’s first-level department data, extracted using the bge-large-en-v1.5 model \citep{bge_embedding}, into a lower-dimensional space. The resulting visualization, depicted in Figure~\ref{fig:distribution}, reveals distinct separations among data clusters, underscoring the robustness and discriminative power of our classification framework.

\section{Training and Inference}

\begin{figure}[t]
\centering
  \includegraphics[width=\textwidth]{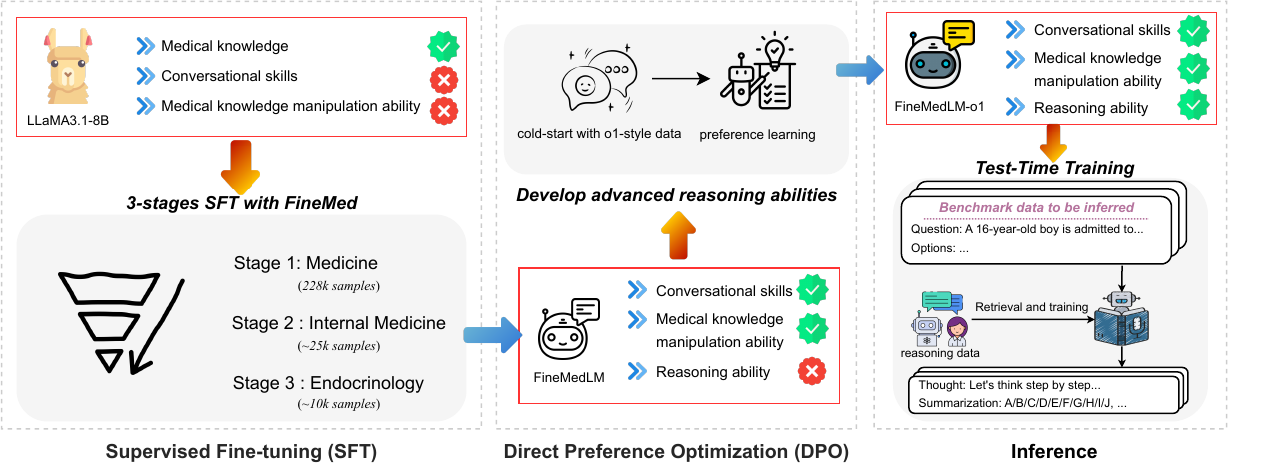}
  \caption{The overall pipeline for training and inference of FineMedLM-o1.}
  \label{fig:train}
\end{figure}

In this section, we detail the two primary stages of model training: SFT and DPO. To further enhance the reasoning capabilities of the medical LLM, we also incorporate TTT during inference. The overall workflow is illustrated in Figure~\ref{fig:train}.

\subsection{FineMedLM}

To enhance the language model’s capability for engaging in natural conversations, we first conduct 3-stage SFT, which involves fine-tuning the pre-trained LLM on chat-style data, including both instructions and responses.

Many researchers \citep{wang2021survey, zhang2024curriculum} have found that models perform better when they are trained incrementally from simple tasks to more complex ones. Based on this principle, multi-stage learning methods have gained increasing popularity. For instance, some researchers employ a multi-stage training approach for continuous pre-training \citep{zhao2024aqulia}. In their method, the first stage involves mixed training with general and medical data, while the second stage focuses solely on medical data. This approach aims to mitigate the significant performance degradation caused by discrepancies between pre-training and fine-tuning datasets, thereby enhancing the model's capabilities in the medical domain. Inspired by this, we adopt a 3-stage SFT approach within medical knowledge, starting broadly with general medical domain data and progressively narrowing the scope to focus on increasingly specialized subfields.

Our model is based on Llama3.1-8B and fine-tuned in 3 stages. In the first stage, we randomly sample 228,000 instances from the entire medical dataset for training, using a learning rate of 1e-5. We reserve 10\% of the training set for validation and obtain the best checkpoint after two epochs. In the second stage, 25,600 instances are randomly selected from the Internal Medicine subset of FineMed, with a learning rate of 5e-6, resulting in the best checkpoint after one epoch. In the third stage, 10,240 instances are selected from the Endocrinology subset, using a learning rate of 5e-6, and the best checkpoint is obtained after one epoch.
The training hyperparameters are as follows: sequence length is set to 1024, batch size to 256, and warm-up steps to 10\% of the total steps. A cosine learning rate scheduler is employed. To mitigate overfitting, we apply a weight decay of 0.01 and set dropout to 0.1. The training stage
 is parallelized across 4 NVIDIA A100-80G GPUs, utilizing the AdamW optimizer with bf16 precision and the ZeRO-3 optimization strategy.

\subsection{FineMedLM-o1}

We obtain FineMedLM by conducting a 3-stage SFT on Llama3.1-8B using carefully curated non-reasoning data. It is a model equipped with both dialogue abilities and medical knowledge but lacks deep reasoning capabilities. To develop FineMedLM-o1 with advanced reasoning capabilities, we apply DPO to FineMedLM. We first utilize dialogue data containing explicit reasoning traces to cold-start the model, thereby equipping it with preliminary reasoning capabilities. Subsequently, for each complex instruction, we perform preference learning using both the verified correct responses (which include reasoning processes) generated by QwQ and the verified incorrect responses generated by the cold-started model, in order to further enhance its ability to reason effectively.

During the cold-start stage, we randomly select 12,800 samples from the FineMed subset for training, which includes complex instructions and long-form reasoning responses. The hyperparameters are set as follows: a learning rate of 5e-6, a batch size of 128, a sequence length of 8192, and 2 epochs. All other hyperparameters are kept consistent with those used for FineMedLM. In the preference learning stage, we utilize all 33,000 samples to construct the DPO dataset and train the model for one epoch with a learning rate of 1e-7, while keeping all other hyperparameters consistent with those used in the cold-start stage.

\subsection{Test-Time Training}

In order to further advance the reasoning capabilities of FineMedLM-o1, we incorporate TTT into the inference process. Some researchers demonstrate that TTT can substantially enhance a model’s reasoning abilities \citep{akyürek2024surprisingeffectivenesstesttimetraining, hübotter2025efficientlylearningtesttimeactive}. Specifically, their findings reveal that incorporating TTT can enhance performance by six times on the Abstraction and Reasoning Corpus (ARC) \citep{chollet2019measureintelligence}, a challenging benchmark for assessing reasoning abilities, compared to a baseline fine-tuned model.

When making inferences on a benchmark dataset, we first use bge-large-en-v1.5 to retrieve the most similar instance from the long-form reasoning subset of FineMed. The model then undergoes training on the retrieved data using the same hyperparameters as FineMedLM-o1. Once training is completed, the model generates an answer for the benchmark instance, after which the model parameters are restored to their original state. This approach ensures that the model leverages relevant contextual knowledge while maintaining its generalization ability. By reverting the parameters, we prevent catastrophic forgetting and preserve the model’s original capabilities for subsequent tasks.

\section{Experiment}

\subsection{Evaluation Setup}

We evaluate the performance of our model on several Chinese and English benchmarks in the medical field. These benchmarks span multiple domains, including biology and healthcare, and encompass both common medical questions and challenging problems that demand complex reasoning. This comprehensive evaluation assesses the model's ability to understand medical knowledge and provide accurate answers to medical queries.

\paragraph{Benchmarks}
We evaluate the models using medical question subsets from the MMLU \citep{hendryckstest2021} and C-Eval \citep{huang2024c} benchmarks, along with questions from the CMB-Exam \citep{wang2023cmb}, CMExam \citep{liu2023benchmarkinglargelanguagemodels}, MedQA \citep{jin2021disease}, and MedMCQA \citep{pal2022medmcqa} test sets to assess their proficiency in medical knowledge. To more clearly illustrate the impact of long-form reasoning data on enhancing the model's reasoning capabilities through DPO and TTT, we incorporate additional medical subsets from the MMLU-Pro benchmark \citep{wang2024mmluprorobustchallengingmultitask} for evaluation. These subsets feature more challenging tasks that require complex reasoning, thereby providing a robust assessment of the model's advanced reasoning ability. Among these benchmarks, C-Eval, CMB-Exam, and CMExam consist of Chinese questions, while the others are in English.

\paragraph{Baselines}
For common medical benchmarks, we compare the performance of our models against both general models and medical fine-tuned models with comparable parameter sizes. The models in our comparison include Baichuan2-7B \citep{baichuan2023baichuan2}, ChatGLM3-6B \citep{glm2024chatglm}, InternLM-7B \citep{cai2024internlm2}, Llama3.1-8B \citep{dubey2024llama}, HuatuoGPT2-7B \citep{chen2023huatuogpt}, and Medical-Llama3-8B\footnote{\url{https://huggingface.co/ruslanmv/Medical-Llama3-8B}}. For benchmarks that require complex reasoning, we compare with the medical reasoning model HuatuoGPT-o1-8B \citep{chen2024huatuogpto1medicalcomplexreasoning}. Additionally, we report experimental results for QwQ-32B-Preview \citep{team2024qwq}, GPT-4o-mini \citep{openai_gpt4o}, GPT-4o, and the recently released DeepSeek-v3 \citep{deepseekai2024deepseekv3technicalreport} as supplementary references.

\subsection{Results} \label{sec:results}

In the experiments, we randomly select three data points from the benchmark training set to perform 3-shot evaluations, ensuring consistency in the model output format. To mitigate potential bias, the experiment is repeated three times, and the average value of the results is reported as the final outcome.

Table~\ref{tab:result1} summarizes the overall performance of various models on standard medical benchmarks. Notably, some newer models (e.g., Llama3.1-8B), including general-purpose models, occasionally surpass specialized medical fine-tuned models on certain benchmarks. Our model, FineMedLM, achieves significant improvements across all benchmarks compared to its base model, Llama3.1-8B, with an average performance gain of 12\%. However, FineMedLM underperforms on Chinese benchmarks (C-Eval, CMB-Exam, CMExam) relative to Baichuan2-7B, which benefits from pretraining on extensive Chinese datasets. FineMedLM-o1 exhibits strong performance across all benchmarks, outperforming FineMedLM by an average of 10\%, highlighting the critical role of robust reasoning capabilities in addressing medical problems.

\begin{table}[t]
  \centering
  \resizebox{\textwidth}{!}{%
    \begin{tabular}{lcccccc}
      \hline
      \textbf{Model}           & \textbf{C-Eval} & \textbf{CMB-Exam} & \textbf{CMExam} & \textbf{MMLU} & \textbf{MedQA} & \textbf{MedMCQA}\\
      \hline
      Baichuan2-7B\citep{baichuan2023baichuan2}        & \underline{56.96} & \underline{51.13} & \underline{50.70} & 50.14 & 41.56 & 41.12  \\
      ChatGLM3-6B\citep{glm2024chatglm}         & 37.97 & 42.94 & 42.01 & 46.74 & 33.78 & 33.71  \\
      InternLM2.5-7B\citep{cai2024internlm2}         & 53.16 & 48.09 & 49.42 & 60.61 & 45.01 & 40.08  \\
      HuatuoGPT2-7B \ding{171}\citep{chen2023huatuogpt} & 51.90 & 45.79 & 45.29 & 45.64 & 40.38 & 34.93  \\
      Medical-Llama3-8B \ding{171}  & 40.51 & 35.09 & 34.87 & 65.47 & 50.20 & 50.68  \\
      Llama3.1-8B\citep{dubey2024llama}         & 43.04 & 40.80 & 41.83 & 67.03 & 55.30 & 53.77  \\
      FineMedLM         \ding{171}   & 55.70 & 48.47 & 46.32 & \underline{71.44} & \underline{57.66} &  \underline{54.91} \\
      FineMedLM-o1      \ding{171}   &  \textbf{62.03} &  \textbf{54.74} &  \textbf{51.93} &  \textbf{78.24} &  \textbf{58.52} &  \textbf{62.75}  \\
      \hdashline 
      FineMedLM-o1 (TTT) \ding{171} & \textbf{65.75} & \textbf{58.02} & \textbf{55.04} & \textbf{81.36} & \textbf{60.86} & \textbf{65.26}  \\
      \hline
    \end{tabular}
  }
  \caption{
    Main results on medical benchmarks. \ding{171} means this LLM is specifically fine-tuned and optimized for tasks within the medical domain. Within each segment, \textbf{bold} highlights the best scores, and \underline{underlines} indicate the second-best.
  }
  \label{tab:result1}
\end{table}

Table~\ref{tab:result2} presents our performance on challenging medical benchmarks that require complex reasoning. FineMedLM-o1 shows a significant improvement in reasoning ability over FineMedLM, with a gain of approximately 27\%.Notably, FineMedLM-o1 achieves superior average performance compared to the recently released HuatuoGPT-o1 on the medical subset of MMLU-Pro. This phenomenon can be attributed to our carefully crafted synthetic data and optimized training process. Furthermore, the introduction of TTT further boosts FineMedLM-o1's reasoning capabilities, bringing its performance on par with GPT-4o-mini.

\begin{table}[t]
  \centering
  \begin{tabular}{lcccc}
    \hline
    \textbf{Model} & \textbf{Size} & \textbf{Biology} & \textbf{Health} & \textbf{Average} \\
    \hline
    FineMedLM          & 8B & 58.72 & 42.42 & 50.57 \\
    HuatuoGPT-o1\citep{huang2024opencoderopencookbooktoptier}       & 8B & 68.20 & \underline{58.70} & 63.45 \\
    FineMedLM-o1       & 8B & \underline{70.71} & 57.95 & \underline{64.33} \\
    FineMedLM-o1 (TTT) & 8B & \textbf{80.54} &  \textbf{66.71} &  \textbf{72.83} \\
    \hdashline
    GPT-4o-mini\citep{openai_gpt4o} & - & 80.20 & 67.60 & 73.90  \\
    QwQ-32B-Preview\citep{team2024qwq} & 32B & 84.10 & 70.66 & 77.38  \\
    GPT-4o\citep{openai_gpt4o}      & - & \underline{86.75} & \underline{72.12} & \underline{79.44}  \\
    DeepSeek-v3\citep{deepseekai2024deepseekv3technicalreport}     & 671B & \textbf{88.15} & \textbf{74.82} & \textbf{81.49}  \\
    \hline
  \end{tabular}
  \caption{
    Main results on MMLU-Pro. \textbf{bold} highlights the best scores, and \underline{underlines} indicate the second-best.
  }
  \label{tab:result2}
\end{table}

\subsection{Ablation Studies}

In this section, we conduct four studies: 
(1) evaluating the effectiveness of our 3-stage SFT process, 
(2) verifying the consistency between our LLM-as-a-judge framework and expert assessments of instruction quality and complexity, 
(3) analyzing the impact of long-form reasoning data on TTT, 
and (4) comparing the additional inference time introduced by TTT.

\paragraph{Ablation for 3-stage SFT}
To assess the effectiveness of the proposed 3-stage SFT approach, we design two baseline methods for comparison. The first baseline involves training solely on FineMed's medical dataset without incorporating multiple stages. The second baseline employs a 3-stage SFT process but with the stages reversed relative to FineMedLM. The results, summarized in Table~\ref{tab:ablation}, demonstrate that FineMedLM outperforms the baseline across all benchmarks, achieving a maximum performance improvement of 15\%. To further demonstrate the effectiveness of the proposed 3-stage SFT framework, we conduct additional experiments on a private benchmark derived from a hospital dataset, with a focus on internal medicine and endocrinology. We evaluate models trained at each stage of the SFT process alongside several strong open-source baselines. A summary of the results is provided in Appendix~\ref{sec:appendixg}. These findings underscore the significant contribution of the 3-stage SFT framework in enhancing the model's ability to effectively encode and utilize medical knowledge.

\begin{table}[t] 
  \centering
  \begin{tabular}{lcccccc}
    \hline
    \textbf{Strategy}           & \textbf{C-Eval} & \textbf{CMB-Exam} & \textbf{CMExam} & \textbf{MMLU} & \textbf{MedQA} & \textbf{MedMCQA}\\
    \hline
    Direct        & 49.37 & 48.04 & 45.69 & 70.33 & 55.54 & 54.48  \\
    Reversed      & 48.10 & 47.95 & 45.42 & 70.06 & 56.33 & 52.69  \\
    FineMedLM     &  \textbf{55.70} &  \textbf{48.47} & \textbf{46.32} &  \textbf{71.44} &  \textbf{57.66} &  \textbf{54.91}  \\
    \hline
  \end{tabular}
  \caption{
Results of ablation experiments for 3-stage SFT. "Direct" means training directly with medical data, and "Reversed" means training by reversing the order of the SFT stages.
  }
  \label{tab:ablation}
\end{table}

\paragraph{Expert Evaluation Comparison}
To assess the consistency of our LLM-as-a-judge approach with expert evaluations, we randomly sample 3,000 instances from the FineMed dataset and invite professional clinicians to evaluate them using the same criteria applied by the LLM. The evaluation results, along with the original LLM scores for comparison, are presented in Table~\ref{tab:expert}. Notably, we observe a high degree of agreement between human and LLM assessments, supporting the reliability of our automatic evaluation method.

\renewcommand{\arraystretch}{1.2} 
\begin{table}[ht]
  \centering
  \resizebox{\textwidth}{!}{%
    \begin{tabular}{lcc}
      \hline
      \textbf{Evaluator}           & \textbf{Quality (Average Score)} & \textbf{Complexity (Average Score)} \\
      \hline
      Expert & 8.35 & 6.33   \\
      Qwen2.5-72B-Instruct\citep{yang2024qwen2}  & 8.27 & 6.41  \\
      \hline
    \end{tabular}
  }
  \caption{
    Main results on expert quality evaluation. We randomly sample 3,000 instances from the FineMed dataset and invite professional physicians to evaluate them using the same criteria as in the LLM-based assessment. For comparison, we also include the original LLM-based scores.
  }
  \label{tab:expert}
\end{table}


\begin{wraptable}{r}{0.55\textwidth} 
\vspace{-23pt}  
  \centering
  \scalebox{0.8}{\begin{tabular}{lcc}
    \hline
    \textbf{Method} & \textbf{Biology} & \textbf{Health} \\
    \hline
    No TTT & 70.71 &  57.95  \\
    TTT with common data & 72.83 &  59.79  \\
    TTT with reasoning data & \textbf{80.54} &  \textbf{66.71}  \\
    \hline
     \end{tabular}}
     \caption{Results of ablation experiments for TTT.}
     \label{tab:ablation2}
 \end{wraptable}

\paragraph{Ablation for long-form reasoning data}
To evaluate the impact of long-form reasoning data on TTT, we compare its performance using common data versus reasoning data. As shown in Table~\ref{tab:ablation2}, TTT with reasoning data outperforms TTT with common data across all benchmarks, achieving an average improvement of 11\%. These results underscore the significant contribution of long-form reasoning data in enhancing TTT's effectiveness. Moreover, comparing the performance changes after introducing TTT in Table~\ref{tab:result1} and Table~\ref{tab:result2} indicates that TTT yields even greater benefits for complex problems that require deep thinking.

\paragraph{Inference Time Comparison}
To better examine the trade-off between inference efficiency and performance gains, we evaluate three models: a non-thinking conversation model, a thinking model, and a thinking model incorporating TTT. The comparison focuses on their inference time and corresponding reasoning behavior, as shown in Table~\ref{tab:timecompare}. From FineMedLM to FineMedLM-o1, the inference time increases by approximately 4.7×, accompanied by the emergence of thinking behaviors. With the introduction of TTT, inference time further increases by 2.3×, while the model exhibits more salient reasoning behaviors. We present a case study that qualitatively compares the outputs of the three models mentioned above. The example, drawn from the MMLU-Pro benchmark, involves a complex inference task and is detailed in Appendix~\ref{sec:appendixf}.

\begin{table}[t]
  \centering
  \begin{tabular}{lcccc}
    \hline
    \textbf{Model} & \textbf{Inference Time} & \textbf{Reasoning Behavior}  \\
    \hline
    FineMedLM          & 3.3s & Absent  \\
    FineMedLM-o1       & 15.4s & Present  \\
    FineMedLM-o1 with TTT   & 35.1s & Enhanced  \\

    \hline
  \end{tabular}
  \caption{
    Comparison of the average inference time per instance and corresponding reasoning behavior among FineMedLM, FineMedLM-o1, and FineMedLM-o1 with TTT.
  }
  \label{tab:timecompare}
\end{table}

\section{Related Work}

The development of medical LLMs is of profound significance to human society. These models can assist doctors in making rapid and accurate diagnoses, formulating treatment plans by integrating extensive medical data and clinical cases, and even supporting medical institutions in optimizing resource allocation to enhance the efficiency and quality of healthcare services \citep{thirunavukarasu2023large, wang2024survey}. Consequently, researchers have pursued diverse approaches to building powerful medical LLMs. For instance, ChatDoctor \citep{li2023chatdoctor} utilizes patient-doctor conversation data and is fine-tuned on Llama to improve language model accuracy in healthcare applications. DISC-MedLLM \citep{bao2023disc} employs a two-stage SFT strategy: the first stage integrates medical domain knowledge and conversational capabilities into the model using large-scale instruction datasets, while the second stage fine-tunes it on a smaller, high-quality dataset curated with the assistance of ChatGPT and filtered by human experts to align with human preferences. 

Despite these advancements, existing models often struggle to deliver professional responses to domain-specific medical questions. To this end, some researchers propose PediatricsGPT \citep{yang2024pediatricsgpt}, a model tailored to pediatrics, which leverages high-quality instruction datasets and intricate training procedures, contributing significantly to pediatric expertise in LLMs. However, these models struggle to solve complex medical problems through step-by-step reasoning and self-correction. Our work builds upon the paradigm of using medical knowledge to perform SFT and DPO on LLMs, introducing innovations in data utilization and training strategies to enhance reasoning capabilities for complex medical problems. Moreover, we address a critical gap in domain-specific medical datasets by releasing high-quality SFT datasets encompassing 5 primary medical specialties and 29 subspecialties, providing a valuable resource to the community. Notably, shortly before the publication of our study, HuatuoGPT-o1 \citep{chen2024huatuogpto1medicalcomplexreasoning} was launched. This medical reasoning model, trained using SFT and Proximal Policy Optimization (PPO) \citep{schulman2017proximal}, provides valuable insights for enhancing the complex medical reasoning capabilities of LLMs. We believe that our research not only advances the field of medical LLMs but also provides valuable methodologies and resources to inspire progress in enhancing LLM reasoning abilities across medicine and other specialized domains.

\section{Conclusion}

In this paper, we propose FineMedLM-o1, a LLM with strong medical reasoning capabilities, designed to address the challenges of complex medical problems by performing SFT and DPO with carefully designed synthetic data. Our novel synthetic dataset construction, training process, and introduction of TTT during inference significantly improve the model's ability to handle deep reasoning on complex medical problems. The outstanding performance of FineMedLM-o1 on various benchmarks validates the effectiveness of our approach. By open-sourcing our dataset and training process, we aim to promote the advancement of complex reasoning capabilities of medical LLMs in the research community.

\clearpage
\section*{Acknowledgments}
This work was supported by the National Natural Science Foundation of China (No. 62172101), and the Science and Technology Commission of Shanghai Municipality(No. 23511100602), and Shanghai Municipal Commission of Economy and Informatization, Corpus Construction for Large Language Models in Pediatric Respiratory Diseases (2024-GZL-RGZN-01013)

\bibliography{colm2025_conference}

\begin{thebibliography}{54}
\providecommand{\natexlab}[1]{#1}
\providecommand{\url}[1]{\texttt{#1}}
\expandafter\ifx\csname urlstyle\endcsname\relax
  \providecommand{\doi}[1]{doi: #1}\else
  \providecommand{\doi}{doi: \begingroup \urlstyle{rm}\Url}\fi

\bibitem[Abdin et~al.(2024)Abdin, Aneja, Behl, Bubeck, Eldan, Gunasekar, Harrison, Hewett, Javaheripi, Kauffmann, et~al.]{abdin2024phi}
Marah Abdin, Jyoti Aneja, Harkirat Behl, S{\'e}bastien Bubeck, Ronen Eldan, Suriya Gunasekar, Michael Harrison, Russell~J Hewett, Mojan Javaheripi, Piero Kauffmann, et~al.
\newblock Phi-4 technical report.
\newblock \emph{arXiv preprint arXiv:2412.08905}, 2024.

\bibitem[Achiam et~al.(2023)Achiam, Adler, Agarwal, Ahmad, Akkaya, Aleman, Almeida, Altenschmidt, Altman, Anadkat, et~al.]{achiam2023gpt}
Josh Achiam, Steven Adler, Sandhini Agarwal, Lama Ahmad, Ilge Akkaya, Florencia~Leoni Aleman, Diogo Almeida, Janko Altenschmidt, Sam Altman, Shyamal Anadkat, et~al.
\newblock Gpt-4 technical report.
\newblock \emph{arXiv preprint arXiv:2303.08774}, 2023.

\bibitem[Akyürek et~al.(2024)Akyürek, Damani, Qiu, Guo, Kim, and Andreas]{akyürek2024surprisingeffectivenesstesttimetraining}
Ekin Akyürek, Mehul Damani, Linlu Qiu, Han Guo, Yoon Kim, and Jacob Andreas.
\newblock The surprising effectiveness of test-time training for abstract reasoning, 2024.
\newblock URL \url{https://arxiv.org/abs/2411.07279}.

\bibitem[Allen-Zhu \& Li(2024{\natexlab{a}})Allen-Zhu and Li]{allenzhu2024physicslanguagemodels31}
Zeyuan Allen-Zhu and Yuanzhi Li.
\newblock Physics of language models: Part 3.1, knowledge storage and extraction, 2024{\natexlab{a}}.
\newblock URL \url{https://arxiv.org/abs/2309.14316}.

\bibitem[Allen-Zhu \& Li(2024{\natexlab{b}})Allen-Zhu and Li]{allenzhu2024physicslanguagemodels32}
Zeyuan Allen-Zhu and Yuanzhi Li.
\newblock Physics of language models: Part 3.2, knowledge manipulation, 2024{\natexlab{b}}.
\newblock URL \url{https://arxiv.org/abs/2309.14402}.

\bibitem[Allen-Zhu \& Li(2024{\natexlab{c}})Allen-Zhu and Li]{allenzhu2024physicslanguagemodels33}
Zeyuan Allen-Zhu and Yuanzhi Li.
\newblock Physics of language models: Part 3.3, knowledge capacity scaling laws, 2024{\natexlab{c}}.
\newblock URL \url{https://arxiv.org/abs/2404.05405}.

\bibitem[Baichuan(2023)]{baichuan2023baichuan2}
Baichuan.
\newblock Baichuan 2: Open large-scale language models.
\newblock \emph{arXiv preprint arXiv:2309.10305}, 2023.
\newblock URL \url{https://arxiv.org/abs/2309.10305}.

\bibitem[Bao et~al.(2023)Bao, Chen, Xiao, Ren, Wu, Zhong, Peng, Huang, and Wei]{bao2023disc}
Zhijie Bao, Wei Chen, Shengze Xiao, Kuang Ren, Jiaao Wu, Cheng Zhong, Jiajie Peng, Xuanjing Huang, and Zhongyu Wei.
\newblock Disc-medllm: Bridging general large language models and real-world medical consultation.
\newblock \emph{arXiv preprint arXiv:2308.14346}, 2023.

\bibitem[Cai et~al.(2024)Cai, Cao, Chen, Chen, Chen, Chen, Chen, Chen, Chen, Chu, Dong, Duan, Fan, Fei, Gao, Ge, Gu, Gu, Gui, Guo, Guo, He, Hu, Huang, Jiang, Jiao, Jin, Lei, Li, Li, Li, Li, Li, Li, Liu, Liu, Hong, Liu, Liu, Liu, Lv, Lv, Lv, Ma, Ma, Ma, Ning, Ouyang, Qiu, Qu, Shang, Shao, Song, Song, Sui, Sun, Sun, Tang, Wang, Wang, Wang, Wang, Wang, Wang, Wang, Wei, Weng, Wu, Xiong, Xu, Xu, Yan, Yan, Yang, Ye, Ying, Yu, Yu, Zang, Zhang, Zhang, Zhang, Zhang, Zhang, Zhang, Zhang, Zhang, Zhang, Zhang, Zhang, Zhao, Zhao, Zhao, Zhou, Zhou, Zhuo, Zou, Qiu, Qiao, and Lin]{cai2024internlm2}
Zheng Cai, Maosong Cao, Haojiong Chen, Kai Chen, Keyu Chen, Xin Chen, Xun Chen, Zehui Chen, Zhi Chen, Pei Chu, Xiaoyi Dong, Haodong Duan, Qi~Fan, Zhaoye Fei, Yang Gao, Jiaye Ge, Chenya Gu, Yuzhe Gu, Tao Gui, Aijia Guo, Qipeng Guo, Conghui He, Yingfan Hu, Ting Huang, Tao Jiang, Penglong Jiao, Zhenjiang Jin, Zhikai Lei, Jiaxing Li, Jingwen Li, Linyang Li, Shuaibin Li, Wei Li, Yining Li, Hongwei Liu, Jiangning Liu, Jiawei Hong, Kaiwen Liu, Kuikun Liu, Xiaoran Liu, Chengqi Lv, Haijun Lv, Kai Lv, Li~Ma, Runyuan Ma, Zerun Ma, Wenchang Ning, Linke Ouyang, Jiantao Qiu, Yuan Qu, Fukai Shang, Yunfan Shao, Demin Song, Zifan Song, Zhihao Sui, Peng Sun, Yu~Sun, Huanze Tang, Bin Wang, Guoteng Wang, Jiaqi Wang, Jiayu Wang, Rui Wang, Yudong Wang, Ziyi Wang, Xingjian Wei, Qizhen Weng, Fan Wu, Yingtong Xiong, Chao Xu, Ruiliang Xu, Hang Yan, Yirong Yan, Xiaogui Yang, Haochen Ye, Huaiyuan Ying, Jia Yu, Jing Yu, Yuhang Zang, Chuyu Zhang, Li~Zhang, Pan Zhang, Peng Zhang, Ruijie Zhang, Shuo Zhang, Songyang Zhang, Wenjian Zhang,
  Wenwei Zhang, Xingcheng Zhang, Xinyue Zhang, Hui Zhao, Qian Zhao, Xiaomeng Zhao, Fengzhe Zhou, Zaida Zhou, Jingming Zhuo, Yicheng Zou, Xipeng Qiu, Yu~Qiao, and Dahua Lin.
\newblock Internlm2 technical report, 2024.

\bibitem[Cao et~al.(2023)Cao, Kang, Wang, and Sun]{cao2023instruction}
Yihan Cao, Yanbin Kang, Chi Wang, and Lichao Sun.
\newblock Instruction mining: Instruction data selection for tuning large language models.
\newblock \emph{arXiv preprint arXiv:2307.06290}, 2023.

\bibitem[Chen et~al.(2023)Chen, Wang, Ji, Gao, Jiang, Chen, Zhang, Song, Xie, Kong, et~al.]{chen2023huatuogpt}
Junying Chen, Xidong Wang, Ke~Ji, Anningzhe Gao, Feng Jiang, Shunian Chen, Hongbo Zhang, Dingjie Song, Wenya Xie, Chuyi Kong, et~al.
\newblock Huatuogpt-ii, one-stage training for medical adaption of llms.
\newblock \emph{arXiv preprint arXiv:2311.09774}, 2023.

\bibitem[Chen et~al.(2024)Chen, Cai, Ji, Wang, Liu, Wang, Hou, and Wang]{chen2024huatuogpto1medicalcomplexreasoning}
Junying Chen, Zhenyang Cai, Ke~Ji, Xidong Wang, Wanlong Liu, Rongsheng Wang, Jianye Hou, and Benyou Wang.
\newblock Huatuogpt-o1, towards medical complex reasoning with llms, 2024.
\newblock URL \url{https://arxiv.org/abs/2412.18925}.

\bibitem[Chollet(2019)]{chollet2019measureintelligence}
François Chollet.
\newblock On the measure of intelligence, 2019.
\newblock URL \url{https://arxiv.org/abs/1911.01547}.

\bibitem[DeepSeek-AI et~al.(2024)DeepSeek-AI, Liu, Feng, Xue, Wang, Wu, Lu, Zhao, Deng, Zhang, Ruan, Dai, Guo, Yang, Chen, Ji, Li, Lin, Dai, Luo, Hao, Chen, Li, Zhang, Bao, Xu, Wang, Zhang, Ding, Xin, Gao, Li, Qu, Cai, Liang, Guo, Ni, Li, Wang, Chen, Chen, Yuan, Qiu, Li, Song, Dong, Hu, Gao, Guan, Huang, Yu, Wang, Zhang, Xu, Xia, Zhao, Wang, Zhang, Li, Wang, Zhang, Zhang, Tang, Li, Tian, Huang, Wang, Zhang, Wang, Zhu, Chen, Du, Chen, Jin, Ge, Zhang, Pan, Wang, Xu, Zhang, Chen, Li, Lu, Zhou, Chen, Wu, Ye, Ye, Ma, Wang, Zhou, Yu, Zhou, Pan, Wang, Yun, Pei, Sun, Xiao, Zeng, Zhao, An, Liu, Liang, Gao, Yu, Zhang, Li, Jin, Wang, Bi, Liu, Wang, Shen, Chen, Zhang, Chen, Nie, Sun, Wang, Cheng, Liu, Xie, Liu, Yu, Song, Shan, Zhou, Yang, Li, Su, Lin, Li, Wang, Wei, Zhu, Zhang, Xu, Xu, Huang, Li, Zhao, Sun, Li, Wang, Yu, Zheng, Zhang, Shi, Xiong, He, Tang, Piao, Wang, Tan, Ma, Liu, Guo, Wu, Ou, Zhu, Wang, Gong, Zou, He, Zha, Xiong, Ma, Yan, Luo, You, Liu, Zhou, Wu, Ren, Ren, Sha, Fu, Xu, Huang, Zhang, Xie, Zhang, Hao,
  Gou, Ma, Yan, Shao, Xu, Wu, Zhang, Li, Gu, Zhu, Liu, Li, Xie, Song, Gao, and Pan]{deepseekai2024deepseekv3technicalreport}
DeepSeek-AI, Aixin Liu, Bei Feng, Bing Xue, Bingxuan Wang, Bochao Wu, Chengda Lu, Chenggang Zhao, Chengqi Deng, Chenyu Zhang, Chong Ruan, Damai Dai, Daya Guo, Dejian Yang, Deli Chen, Dongjie Ji, Erhang Li, Fangyun Lin, Fucong Dai, Fuli Luo, Guangbo Hao, Guanting Chen, Guowei Li, H.~Zhang, Han Bao, Hanwei Xu, Haocheng Wang, Haowei Zhang, Honghui Ding, Huajian Xin, Huazuo Gao, Hui Li, Hui Qu, J.~L. Cai, Jian Liang, Jianzhong Guo, Jiaqi Ni, Jiashi Li, Jiawei Wang, Jin Chen, Jingchang Chen, Jingyang Yuan, Junjie Qiu, Junlong Li, Junxiao Song, Kai Dong, Kai Hu, Kaige Gao, Kang Guan, Kexin Huang, Kuai Yu, Lean Wang, Lecong Zhang, Lei Xu, Leyi Xia, Liang Zhao, Litong Wang, Liyue Zhang, Meng Li, Miaojun Wang, Mingchuan Zhang, Minghua Zhang, Minghui Tang, Mingming Li, Ning Tian, Panpan Huang, Peiyi Wang, Peng Zhang, Qiancheng Wang, Qihao Zhu, Qinyu Chen, Qiushi Du, R.~J. Chen, R.~L. Jin, Ruiqi Ge, Ruisong Zhang, Ruizhe Pan, Runji Wang, Runxin Xu, Ruoyu Zhang, Ruyi Chen, S.~S. Li, Shanghao Lu, Shangyan Zhou, Shanhuang
  Chen, Shaoqing Wu, Shengfeng Ye, Shengfeng Ye, Shirong Ma, Shiyu Wang, Shuang Zhou, Shuiping Yu, Shunfeng Zhou, Shuting Pan, T.~Wang, Tao Yun, Tian Pei, Tianyu Sun, W.~L. Xiao, Wangding Zeng, Wanjia Zhao, Wei An, Wen Liu, Wenfeng Liang, Wenjun Gao, Wenqin Yu, Wentao Zhang, X.~Q. Li, Xiangyue Jin, Xianzu Wang, Xiao Bi, Xiaodong Liu, Xiaohan Wang, Xiaojin Shen, Xiaokang Chen, Xiaokang Zhang, Xiaosha Chen, Xiaotao Nie, Xiaowen Sun, Xiaoxiang Wang, Xin Cheng, Xin Liu, Xin Xie, Xingchao Liu, Xingkai Yu, Xinnan Song, Xinxia Shan, Xinyi Zhou, Xinyu Yang, Xinyuan Li, Xuecheng Su, Xuheng Lin, Y.~K. Li, Y.~Q. Wang, Y.~X. Wei, Y.~X. Zhu, Yang Zhang, Yanhong Xu, Yanhong Xu, Yanping Huang, Yao Li, Yao Zhao, Yaofeng Sun, Yaohui Li, Yaohui Wang, Yi~Yu, Yi~Zheng, Yichao Zhang, Yifan Shi, Yiliang Xiong, Ying He, Ying Tang, Yishi Piao, Yisong Wang, Yixuan Tan, Yiyang Ma, Yiyuan Liu, Yongqiang Guo, Yu~Wu, Yuan Ou, Yuchen Zhu, Yuduan Wang, Yue Gong, Yuheng Zou, Yujia He, Yukun Zha, Yunfan Xiong, Yunxian Ma, Yuting Yan, Yuxiang
  Luo, Yuxiang You, Yuxuan Liu, Yuyang Zhou, Z.~F. Wu, Z.~Z. Ren, Zehui Ren, Zhangli Sha, Zhe Fu, Zhean Xu, Zhen Huang, Zhen Zhang, Zhenda Xie, Zhengyan Zhang, Zhewen Hao, Zhibin Gou, Zhicheng Ma, Zhigang Yan, Zhihong Shao, Zhipeng Xu, Zhiyu Wu, Zhongyu Zhang, Zhuoshu Li, Zihui Gu, Zijia Zhu, Zijun Liu, Zilin Li, Ziwei Xie, Ziyang Song, Ziyi Gao, and Zizheng Pan.
\newblock Deepseek-v3 technical report, 2024.
\newblock URL \url{https://arxiv.org/abs/2412.19437}.

\bibitem[Dong et~al.(2024)Dong, Li, Dai, Zheng, Ma, Li, Xia, Xu, Wu, Chang, et~al.]{dong2024survey}
Qingxiu Dong, Lei Li, Damai Dai, Ce~Zheng, Jingyuan Ma, Rui Li, Heming Xia, Jingjing Xu, Zhiyong Wu, Baobao Chang, et~al.
\newblock A survey on in-context learning.
\newblock In \emph{Proceedings of the 2024 Conference on Empirical Methods in Natural Language Processing}, pp.\  1107--1128, 2024.

\bibitem[Dubey et~al.(2024)Dubey, Jauhri, Pandey, Kadian, Al-Dahle, Letman, Mathur, Schelten, Yang, Fan, et~al.]{dubey2024llama}
Abhimanyu Dubey, Abhinav Jauhri, Abhinav Pandey, Abhishek Kadian, Ahmad Al-Dahle, Aiesha Letman, Akhil Mathur, Alan Schelten, Amy Yang, Angela Fan, et~al.
\newblock The llama 3 herd of models.
\newblock \emph{arXiv preprint arXiv:2407.21783}, 2024.

\bibitem[GLM et~al.(2024)GLM, Zeng, Xu, Wang, Zhang, Yin, Rojas, Feng, Zhao, Lai, Yu, Wang, Sun, Zhang, Cheng, Gui, Tang, Zhang, Li, Zhao, Wu, Zhong, Liu, Huang, Zhang, Zheng, Lu, Duan, Zhang, Cao, Yang, Tam, Zhao, Liu, Xia, Zhang, Gu, Lv, Liu, Liu, Yang, Song, Zhang, An, Xu, Niu, Yang, Li, Bai, Dong, Qi, Wang, Yang, Du, Hou, and Wang]{glm2024chatglm}
Team GLM, Aohan Zeng, Bin Xu, Bowen Wang, Chenhui Zhang, Da~Yin, Diego Rojas, Guanyu Feng, Hanlin Zhao, Hanyu Lai, Hao Yu, Hongning Wang, Jiadai Sun, Jiajie Zhang, Jiale Cheng, Jiayi Gui, Jie Tang, Jing Zhang, Juanzi Li, Lei Zhao, Lindong Wu, Lucen Zhong, Mingdao Liu, Minlie Huang, Peng Zhang, Qinkai Zheng, Rui Lu, Shuaiqi Duan, Shudan Zhang, Shulin Cao, Shuxun Yang, Weng~Lam Tam, Wenyi Zhao, Xiao Liu, Xiao Xia, Xiaohan Zhang, Xiaotao Gu, Xin Lv, Xinghan Liu, Xinyi Liu, Xinyue Yang, Xixuan Song, Xunkai Zhang, Yifan An, Yifan Xu, Yilin Niu, Yuantao Yang, Yueyan Li, Yushi Bai, Yuxiao Dong, Zehan Qi, Zhaoyu Wang, Zhen Yang, Zhengxiao Du, Zhenyu Hou, and Zihan Wang.
\newblock Chatglm: A family of large language models from glm-130b to glm-4 all tools, 2024.

\bibitem[Gururajan et~al.(2024)Gururajan, Lopez-Cuena, Bayarri-Planas, Tormos, Hinjos, Bernabeu-Perez, Arias-Duart, Martin-Torres, Urcelay-Ganzabal, Gonzalez-Mallo, et~al.]{gururajan2024aloe}
Ashwin~Kumar Gururajan, Enrique Lopez-Cuena, Jordi Bayarri-Planas, Adrian Tormos, Daniel Hinjos, Pablo Bernabeu-Perez, Anna Arias-Duart, Pablo~Agustin Martin-Torres, Lucia Urcelay-Ganzabal, Marta Gonzalez-Mallo, et~al.
\newblock Aloe: A family of fine-tuned open healthcare llms.
\newblock \emph{arXiv preprint arXiv:2405.01886}, 2024.

\bibitem[Hendrycks et~al.(2021)Hendrycks, Burns, Basart, Zou, Mazeika, Song, and Steinhardt]{hendryckstest2021}
Dan Hendrycks, Collin Burns, Steven Basart, Andy Zou, Mantas Mazeika, Dawn Song, and Jacob Steinhardt.
\newblock Measuring massive multitask language understanding.
\newblock \emph{Proceedings of the International Conference on Learning Representations (ICLR)}, 2021.

\bibitem[Huang et~al.(2024{\natexlab{a}})Huang, Cheng, Liu, Hao, Song, Xu, Yang, Liu, Zhang, Chai, Yuan, Zhang, Fu, Liu, Zhang, Wang, Qi, Xu, and Chu]{huang2024opencoderopencookbooktoptier}
Siming Huang, Tianhao Cheng, J.~K. Liu, Jiaran Hao, Liuyihan Song, Yang Xu, J.~Yang, J.~H. Liu, Chenchen Zhang, Linzheng Chai, Ruifeng Yuan, Zhaoxiang Zhang, Jie Fu, Qian Liu, Ge~Zhang, Zili Wang, Yuan Qi, Yinghui Xu, and Wei Chu.
\newblock Opencoder: The open cookbook for top-tier code large language models, 2024{\natexlab{a}}.
\newblock URL \url{https://arxiv.org/abs/2411.04905}.

\bibitem[Huang et~al.(2024{\natexlab{b}})Huang, Bai, Zhu, Zhang, Zhang, Su, Liu, Lv, Zhang, Fu, et~al.]{huang2024c}
Yuzhen Huang, Yuzhuo Bai, Zhihao Zhu, Junlei Zhang, Jinghan Zhang, Tangjun Su, Junteng Liu, Chuancheng Lv, Yikai Zhang, Yao Fu, et~al.
\newblock C-eval: A multi-level multi-discipline chinese evaluation suite for foundation models.
\newblock \emph{Advances in Neural Information Processing Systems}, 36, 2024{\natexlab{b}}.

\bibitem[Hübotter et~al.(2025)Hübotter, Bongni, Hakimi, and Krause]{hübotter2025efficientlylearningtesttimeactive}
Jonas Hübotter, Sascha Bongni, Ido Hakimi, and Andreas Krause.
\newblock Efficiently learning at test-time: Active fine-tuning of llms, 2025.
\newblock URL \url{https://arxiv.org/abs/2410.08020}.

\bibitem[Ji et~al.(2025)Ji, Zhao, Wang, Wang, Zhang, Cheng, Feng, and Zhang]{ji-etal-2025-robguard}
Changkai Ji, Bowen Zhao, Zhuoyao Wang, Yingwen Wang, Yuejie Zhang, Ying Cheng, Rui Feng, and Xiaobo Zhang.
\newblock {R}o{BG}uard: Enhancing {LLM}s to assess risk of bias in clinical trial documents.
\newblock In Owen Rambow, Leo Wanner, Marianna Apidianaki, Hend Al-Khalifa, Barbara~Di Eugenio, and Steven Schockaert (eds.), \emph{Proceedings of the 31st International Conference on Computational Linguistics}, pp.\  1258--1277, Abu Dhabi, UAE, January 2025. Association for Computational Linguistics.
\newblock URL \url{https://aclanthology.org/2025.coling-main.84/}.

\bibitem[Jin et~al.(2021)Jin, Pan, Oufattole, Weng, Fang, and Szolovits]{jin2021disease}
Di~Jin, Eileen Pan, Nassim Oufattole, Wei-Hung Weng, Hanyi Fang, and Peter Szolovits.
\newblock What disease does this patient have? a large-scale open domain question answering dataset from medical exams.
\newblock \emph{Applied Sciences}, 11\penalty0 (14):\penalty0 6421, 2021.

\bibitem[Kwon et~al.(2023)Kwon, Li, Zhuang, Sheng, Zheng, Yu, Gonzalez, Zhang, and Stoica]{kwon2023efficientmemorymanagementlarge}
Woosuk Kwon, Zhuohan Li, Siyuan Zhuang, Ying Sheng, Lianmin Zheng, Cody~Hao Yu, Joseph~E. Gonzalez, Hao Zhang, and Ion Stoica.
\newblock Efficient memory management for large language model serving with pagedattention, 2023.
\newblock URL \url{https://arxiv.org/abs/2309.06180}.

\bibitem[Li et~al.(2023)Li, Li, Zhang, Dan, Jiang, and Zhang]{li2023chatdoctor}
Yunxiang Li, Zihan Li, Kai Zhang, Ruilong Dan, Steve Jiang, and You Zhang.
\newblock Chatdoctor: A medical chat model fine-tuned on a large language model meta-ai (llama) using medical domain knowledge.
\newblock \emph{Cureus}, 15\penalty0 (6), 2023.

\bibitem[Liu et~al.(2023{\natexlab{a}})Liu, Zhou, Hua, Chong, Tian, Liu, Wang, You, Guo, Zhu, and Li]{liu2023benchmarkinglargelanguagemodels}
Junling Liu, Peilin Zhou, Yining Hua, Dading Chong, Zhongyu Tian, Andrew Liu, Helin Wang, Chenyu You, Zhenhua Guo, Lei Zhu, and Michael~Lingzhi Li.
\newblock Benchmarking large language models on cmexam -- a comprehensive chinese medical exam dataset, 2023{\natexlab{a}}.
\newblock URL \url{https://arxiv.org/abs/2306.03030}.

\bibitem[Liu et~al.(2023{\natexlab{b}})Liu, Zeng, He, Jiang, and He]{liu2023makes}
Wei Liu, Weihao Zeng, Keqing He, Yong Jiang, and Junxian He.
\newblock What makes good data for alignment? a comprehensive study of automatic data selection in instruction tuning.
\newblock \emph{arXiv preprint arXiv:2312.15685}, 2023{\natexlab{b}}.

\bibitem[Lu et~al.(2023)Lu, Yuan, Yuan, Lin, Lin, Tan, Zhou, and Zhou]{lu2023instag}
Keming Lu, Hongyi Yuan, Zheng Yuan, Runji Lin, Junyang Lin, Chuanqi Tan, Chang Zhou, and Jingren Zhou.
\newblock \# instag: Instruction tagging for analyzing supervised fine-tuning of large language models.
\newblock In \emph{The Twelfth International Conference on Learning Representations}, 2023.

\bibitem[M-A-P(2024)]{finefineweb}
Xinrun Du* Zhimiao Yu* Zili Wang* Zekun Wang Shuyue Guo Tianyu Zheng Kang Zhu Jerry Liu Shawn Yue Binbin Liu Zhongyuan Peng Yifan Yao Jack Yang Ziming Li Bingni Zhang Minghao Liu Tianyu Liu Yang Gao Wenhu Chen Xiaohuan Zhou Qian Liu Taifeng Wang+ Wenhao~Huang+ M-A-P, Ge~Zhang*.
\newblock Finefineweb: A comprehensive study on fine-grained domain web corpus, December 2024.
\newblock URL \url{[https://huggingface.co/datasets/m-a-p/FineFineWeb](https://huggingface.co/datasets/m-a-p/FineFineWeb)}.

\bibitem[OpenAI(2024)]{openai_gpt4o}
OpenAI.
\newblock Hello gpt-4, 2024.
\newblock URL \url{https://openai.com/index/hello-gpt-4o/}.

\bibitem[Pal et~al.(2022)Pal, Umapathi, and Sankarasubbu]{pal2022medmcqa}
Ankit Pal, Logesh~Kumar Umapathi, and Malaikannan Sankarasubbu.
\newblock Medmcqa: A large-scale multi-subject multi-choice dataset for medical domain question answering.
\newblock In \emph{Conference on health, inference, and learning}, pp.\  248--260. PMLR, 2022.

\bibitem[Rafailov et~al.(2024)Rafailov, Sharma, Mitchell, Manning, Ermon, and Finn]{rafailov2024direct}
Rafael Rafailov, Archit Sharma, Eric Mitchell, Christopher~D Manning, Stefano Ermon, and Chelsea Finn.
\newblock Direct preference optimization: Your language model is secretly a reward model.
\newblock \emph{Advances in Neural Information Processing Systems}, 36, 2024.

\bibitem[Schulman et~al.(2017)Schulman, Wolski, Dhariwal, Radford, and Klimov]{schulman2017proximal}
John Schulman, Filip Wolski, Prafulla Dhariwal, Alec Radford, and Oleg Klimov.
\newblock Proximal policy optimization algorithms.
\newblock \emph{arXiv preprint arXiv:1707.06347}, 2017.

\bibitem[Shao et~al.(2024)Shao, Wang, Zhu, Xu, Song, Bi, Zhang, Zhang, Li, Wu, and Guo]{shao2024deepseekmathpushinglimitsmathematical}
Zhihong Shao, Peiyi Wang, Qihao Zhu, Runxin Xu, Junxiao Song, Xiao Bi, Haowei Zhang, Mingchuan Zhang, Y.~K. Li, Y.~Wu, and Daya Guo.
\newblock Deepseekmath: Pushing the limits of mathematical reasoning in open language models, 2024.
\newblock URL \url{https://arxiv.org/abs/2402.03300}.

\bibitem[Team(2024)]{team2024qwq}
Qwen Team.
\newblock Qwq: Reflect deeply on the boundaries of the unknown, 2024.

\bibitem[Thirunavukarasu et~al.(2023)Thirunavukarasu, Ting, Elangovan, Gutierrez, Tan, and Ting]{thirunavukarasu2023large}
Arun~James Thirunavukarasu, Darren Shu~Jeng Ting, Kabilan Elangovan, Laura Gutierrez, Ting~Fang Tan, and Daniel Shu~Wei Ting.
\newblock Large language models in medicine.
\newblock \emph{Nature medicine}, 29\penalty0 (8):\penalty0 1930--1940, 2023.

\bibitem[Tian et~al.(2024{\natexlab{a}})Tian, Huang, Cheng, He, Fang, Feng, Geng, and Zhang]{tian2024medical}
Weiwei Tian, Xinyu Huang, Tianhao Cheng, Wen He, Jinwu Fang, Rui Feng, Daoying Geng, and Xiaobo Zhang.
\newblock A medical multimodal large language model for pediatric pneumonia.
\newblock \emph{arXiv preprint arXiv:2409.02608}, 2024{\natexlab{a}}.

\bibitem[Tian et~al.(2024{\natexlab{b}})Tian, Gan, Song, Zhang, and Zhang]{tian2024chimedgptchinesemedicallarge}
Yuanhe Tian, Ruyi Gan, Yan Song, Jiaxing Zhang, and Yongdong Zhang.
\newblock Chimed-gpt: A chinese medical large language model with full training regime and better alignment to human preferences, 2024{\natexlab{b}}.
\newblock URL \url{https://arxiv.org/abs/2311.06025}.

\bibitem[Wang et~al.(2024{\natexlab{a}})Wang, Ning, Peng, Wei, Tesfai, Mao, Zhu, and Huang]{wang2024survey}
Jinqiang Wang, Huansheng Ning, Yi~Peng, Qikai Wei, Daniel Tesfai, Wenwei Mao, Tao Zhu, and Runhe Huang.
\newblock A survey on large language models from general purpose to medical applications: Datasets, methodologies, and evaluations.
\newblock \emph{arXiv preprint arXiv:2406.10303}, 2024{\natexlab{a}}.

\bibitem[Wang et~al.(2023)Wang, Chen, Song, Zhang, Chen, Xiao, Jiang, Li, Wan, Wang, et~al.]{wang2023cmb}
Xidong Wang, Guiming~Hardy Chen, Dingjie Song, Zhiyi Zhang, Zhihong Chen, Qingying Xiao, Feng Jiang, Jianquan Li, Xiang Wan, Benyou Wang, et~al.
\newblock Cmb: A comprehensive medical benchmark in chinese.
\newblock \emph{arXiv preprint arXiv:2308.08833}, 2023.

\bibitem[Wang et~al.(2021)Wang, Chen, and Zhu]{wang2021survey}
Xin Wang, Yudong Chen, and Wenwu Zhu.
\newblock A survey on curriculum learning.
\newblock \emph{IEEE transactions on pattern analysis and machine intelligence}, 44\penalty0 (9):\penalty0 4555--4576, 2021.

\bibitem[Wang et~al.(2024{\natexlab{b}})Wang, Ma, Zhang, Ni, Chandra, Guo, Ren, Arulraj, He, Jiang, Li, Ku, Wang, Zhuang, Fan, Yue, and Chen]{wang2024mmluprorobustchallengingmultitask}
Yubo Wang, Xueguang Ma, Ge~Zhang, Yuansheng Ni, Abhranil Chandra, Shiguang Guo, Weiming Ren, Aaran Arulraj, Xuan He, Ziyan Jiang, Tianle Li, Max Ku, Kai Wang, Alex Zhuang, Rongqi Fan, Xiang Yue, and Wenhu Chen.
\newblock Mmlu-pro: A more robust and challenging multi-task language understanding benchmark, 2024{\natexlab{b}}.
\newblock URL \url{https://arxiv.org/abs/2406.01574}.

\bibitem[Wu et~al.(2024)Wu, Hasan, Wu, Kim, Cheung, Zhang, and Wu]{wu2024chainofthoughcotpromptingstrategies}
Zhaolong Wu, Abul Hasan, Jinge Wu, Yunsoo Kim, Jason P.~Y. Cheung, Teng Zhang, and Honghan Wu.
\newblock Chain-of-though (cot) prompting strategies for medical error detection and correction, 2024.
\newblock URL \url{https://arxiv.org/abs/2406.09103}.

\bibitem[Xiao et~al.(2023)Xiao, Liu, Zhang, and Muennighoff]{bge_embedding}
Shitao Xiao, Zheng Liu, Peitian Zhang, and Niklas Muennighoff.
\newblock C-pack: Packaged resources to advance general chinese embedding, 2023.

\bibitem[Xiong et~al.(2024)Xiong, Dong, Ye, Wang, Zhong, Ji, Jiang, and Zhang]{xiong2024iterative}
Wei Xiong, Hanze Dong, Chenlu Ye, Ziqi Wang, Han Zhong, Heng Ji, Nan Jiang, and Tong Zhang.
\newblock Iterative preference learning from human feedback: Bridging theory and practice for rlhf under kl-constraint, 2024.

\bibitem[Xu et~al.(2023)Xu, Sun, Zheng, Geng, Zhao, Feng, Tao, and Jiang]{xu2023wizardlm}
Can Xu, Qingfeng Sun, Kai Zheng, Xiubo Geng, Pu~Zhao, Jiazhan Feng, Chongyang Tao, and Daxin Jiang.
\newblock Wizardlm: Empowering large language models to follow complex instructions.
\newblock \emph{arXiv preprint arXiv:2304.12244}, 2023.

\bibitem[Yang et~al.(2024{\natexlab{a}})Yang, Yang, Zhang, Hui, Zheng, Yu, Li, Liu, Huang, Wei, et~al.]{yang2024qwen2}
An~Yang, Baosong Yang, Beichen Zhang, Binyuan Hui, Bo~Zheng, Bowen Yu, Chengyuan Li, Dayiheng Liu, Fei Huang, Haoran Wei, et~al.
\newblock Qwen2. 5 technical report.
\newblock \emph{arXiv preprint arXiv:2412.15115}, 2024{\natexlab{a}}.

\bibitem[Yang et~al.(2024{\natexlab{b}})Yang, Wei, Xiao, Wang, Wu, Li, Li, Wang, Chen, Jiang, et~al.]{yang2024pediatricsgpt}
Dingkang Yang, Jinjie Wei, Dongling Xiao, Shunli Wang, Tong Wu, Gang Li, Mingcheng Li, Shuaibing Wang, Jiawei Chen, Yue Jiang, et~al.
\newblock Pediatricsgpt: Large language models as chinese medical assistants for pediatric applications.
\newblock \emph{Advances in Neural Information Processing Systems}, 37:\penalty0 138632--138662, 2024{\natexlab{b}}.

\bibitem[Yue et~al.(2024)Yue, Zheng, Zhang, and Chen]{yue2024mammoth2}
Xiang Yue, Tuney Zheng, Ge~Zhang, and Wenhu Chen.
\newblock Mammoth2: Scaling instructions from the web.
\newblock \emph{arXiv preprint arXiv:2405.03548}, 2024.

\bibitem[Zhang et~al.(2024{\natexlab{a}})Zhang, Lin, Cheng, Xu, Lu, He, Yu, Peng, Zhang, Zou, et~al.]{zhang2024retfound}
Juzhao Zhang, Senlin Lin, Tianhao Cheng, Yi~Xu, Lina Lu, Jiangnan He, Tao Yu, Yajun Peng, Yuejie Zhang, Haidong Zou, et~al.
\newblock Retfound-enhanced community-based fundus disease screening: real-world evidence and decision curve analysis.
\newblock \emph{npj Digital Medicine}, 7\penalty0 (1):\penalty0 108, 2024{\natexlab{a}}.

\bibitem[Zhang et~al.(2024{\natexlab{b}})Zhang, Wang, and Zhao]{zhang2024curriculum}
Zheng Zhang, Junxiang Wang, and Liang Zhao.
\newblock Curriculum learning for graph neural networks: Which edges should we learn first.
\newblock \emph{Advances in Neural Information Processing Systems}, 36, 2024{\natexlab{b}}.

\bibitem[Zhao et~al.(2024)Zhao, Zeng, Shi, Zhou, Hao, and Lin]{zhao2024aqulia}
Lulu Zhao, Weihao Zeng, Xiaofeng Shi, Hua Zhou, Donglin Hao, and Yonghua Lin.
\newblock Aqulia-med llm: Pioneering full-process open-source medical language models.
\newblock \emph{arXiv preprint arXiv:2406.12182}, 2024.

\bibitem[Zhou et~al.(2024)Zhou, Liu, Xu, Iyer, Sun, Mao, Ma, Efrat, Yu, Yu, et~al.]{zhou2024lima}
Chunting Zhou, Pengfei Liu, Puxin Xu, Srinivasan Iyer, Jiao Sun, Yuning Mao, Xuezhe Ma, Avia Efrat, Ping Yu, Lili Yu, et~al.
\newblock Lima: Less is more for alignment.
\newblock \emph{Advances in Neural Information Processing Systems}, 36, 2024.

\end{thebibliography}
\bibliographystyle{colm2025_conference}

\appendix
\clearpage

\section{Limitations}

Our data generation pipeline is built on seeds from FineFineWeb, which in turn are derived from FineWeb. Both datasets have undergone multiple rounds of deduplication and quality filtering to enhance data diversity and reliability. Nevertheless, additional quality assessment may remain beneficial. Due to limited computational resources, however, we randomly selected raw medical texts during the data selection phase without performing prior quality evaluation.
Additionally, the number of selected texts is relatively small. Recognizing the critical role of diverse and high-quality data in SFT, our future work aims to incorporate more extensive, higher-quality, and larger medical datasets. Furthermore, we are exploring improvements to the DPO stage. In future research, we aspire to develop a reinforcement learning algorithm specifically tailored to the medical domain, enabling more sophisticated analyses and applications in medical scenarios.

\section{Synthetic Data Generation Prompts}
\label{sec:appendixa}

This section provides a comprehensive overview of all the prompts utilized during the pipeline of generating synthetic data.

\subsection{Instruction Generation Prompt}\label{sec:a1}
\begin{table}[ht]

\renewcommand{\arraystretch}{1.1}

\small
\centering
    \begin{tabular}{p{1\textwidth}}
    \toprule
    \textbf{\textsc{Instruction Generation}}\\
    \midrule
    You need to generate two questions based on the given text content. These questions can be open-ended, detail-oriented, or related to a broader discussion of the content, but avoid relying on specific case details from the text. Follow these requirements:\\
    Requirements:\\
    1. Make sure the questions are closely related to the main points or themes mentioned in the text.\\
    2. Ensure the two questions are as diverse as possible, avoiding homogeneity.\\
    3. Ensure the questions include all the information needed for the answers. If necessary, add introductory information to the questions.\\
    4. Avoid repetitive or overly simplistic questions, ensuring diversity and depth.\\
    5. The questions must be self-contained and should not require the provided text as background to be understood.\\
    Please rewrite the following text into related questions, and strictly follow the format below for output:\\
    \{\{\\
        \hspace{2em} "question1": "Generated first question",\\
        \hspace{2em} "question2": "Generated second question"\\
    \}\}\\
    Here is the sample text:\\
    \{text\}\\
    \bottomrule
    \end{tabular}
    \caption{Prompts for generating instructions. The text refers to individual data samples drawn from the medical subset of FineFineWeb.} 
    \label{tab:InsGen}
\end{table}

\clearpage

\subsection{Instruction Scoring Prompt}\label{sec:a2}
\begin{table*}[ht]
\renewcommand{\arraystretch}{1.1}
\small
\centering
    \begin{tabular}{p{1.0\textwidth}}
    \toprule
    \textbf{\textsc{Instruction score generation}}\\
    \midrule
    You need to evaluate the given query based on the following criteria and output the results in JSON format. The output should include three parts: quality, difficulty, and whether additional necessary information is required to answer the query. Please follow the scoring standards below:\\
    1. Quality (Score 1--10): Assess the clarity and accuracy of the query. If the query is a simple statement without any question or instruction, score it 1--2.\\
        \hspace{2em} 9--10: Very clear, accurate expression, no ambiguity.\\
        \hspace{2em} 7--8: Clear, accurate expression, but may have minimal ambiguity.\\
        \hspace{2em} 5--6: Fairly clear, generally accurate expression, but some ambiguity exists.\\
        \hspace{2em} 3--4: Not very clear, somewhat vague expression, with obvious ambiguity.\\
        \hspace{2em} 1--2: Unclear, very vague expression, difficult to understand or a simple statement.\\
    2. Difficulty (Score 1--10): Assess the difficulty of understanding and answering the query.\\
        \hspace{2em} 9--10: Very difficult, requires specialized knowledge and complex analysis to answer.\\
        \hspace{2em} 7--8: Quite difficult, requires some specialized knowledge and analysis.\\
        \hspace{2em} 5--6: Moderate difficulty, requires general knowledge and analysis.\\
        \hspace{2em} 3--4: Fairly simple, can be answered with basic knowledge.\\
        \hspace{2em} 1--2: Very simple, no special knowledge required to answer.\\
    3. Relevance to medicine (1--6): Assess the medical relevance of the query.\\
        \hspace{2em} 5--6: Completely related to medicine, with many medical terms appearing.\\
        \hspace{2em} 3--4: Related to medicine, the content involves medical fields.\\
        \hspace{2em} 1--2: Very weak medical relevance.\\
    4. Mention specific details: Whether specific case details in the text are mentioned.\\
    Please strictly follow the format below for output:\\
    \{\{\\
        \hspace{2em} "quality": 1--10,\\
        \hspace{2em} "difficulty": 1--10,\\
        \hspace{2em} "Relevance2Medicine": 1--6,\\
        \hspace{2em} "MentionSpecificDetails": \texttt{"True"}/\texttt{"False"}\\
    \}\}\\
    Please evaluate the following query:\\
    \{instruction\}\\
    \bottomrule
    \end{tabular}
    \caption{Prompts for generating scores for all instructions. The instruction refers to a single data sample generated by Qwen.} 
    \label{tab:InsScoring}
\end{table*}

\clearpage

\subsection{Response Generation Prompts}\label{sec:a3}
\begin{table*}[ht]

\renewcommand{\arraystretch}{1.1}

\small
\centering
    \begin{tabular}{p{1.0\textwidth}}
    \toprule
    \textbf{\textsc{Qwen Response Generation}}\\
    \midrule
    You need to generate two different styles of answers based on the given question. Use the background information provided in the text to assist in formulating a relevant and detailed answer. Follow these answer guidelines:\\
    1. Ensure the answer is closely related to the main points or themes mentioned in the question.\\
    2. Utilize the text content to provide a comprehensive and accurate answer.\\
    3. Ensure proper formatting and readability, including the correct rendering of any LaTeX or mathematical symbols.\\
    4. Ensure that the aMedClsnswer provides a complete solution or explanation, with clear and detailed steps.\\
    5. Please strictly follow the format below for output:\\
    \{\{\\
        \hspace{2em}"answer1": "Generated first answer content",\\
        \hspace{2em}"answer2": "Generated second answer content"\\
    \}\}\\
    Here is the question: \\
    \{instruction\}\\
    Here is the text:\\
    \{text\}\\
    \midrule
    \textbf{\textsc{QwQ Response Generation}}\\
    \midrule
    You need to generate an answer based on the given problem and thoroughly explore the problem through a systematic and long-term thinking process to provide a final and accurate solution. This requires a comprehensive cycle of analysis, summary, exploration, re-evaluation, reflection, backtracking and iteration to form a thoughtful thinking process. Use the background information provided in the text to assist in formulating the answer. Follow these answer guidelines:\\
    1. Please structure your response into two main sections: Thought and Summarization.\\
    2. During the thinking phase, think step by step based on the given text content. If the text content is used, it must be expressed.\\
    3. During the summary phase, based on the thinking process in the thinking phase, give the final answer to the question.\\
    Here is the question: \\
    \{instruction\}\\
    Here is the text:\\
    \{text\}\\
    \midrule
    \textbf{\textsc{Response Verification}}\\
    \midrule
    You need to determine whether the model's answer is correct, based on the given instruction and the original text that contains both the question (as part of the instruction) and its corresponding answer. Please first think carefully and analyze before making a final judgment.\\
    Evaluate the model's answer from the following two perspectives:\\
    1. Instruction Following: Does the answer directly respond to the question in the instruction? If the content of the answer is irrelevant or unrelated to the instruction, it should be considered incorrect.\\
    2. Answer Grounding: Is the answer supported by the original text? If any part of the answer cannot be verified or retrieved from the original text, it should be considered incorrect.\\
    If both conditions are met, the answer is correct. At the end, respond with only "Yes" or "No".\\
    Here is the instruction: \\
    \{instruction\}\\
    Here is the original text:\\
    \{text\}\\
    Here is the model's answer:\\
    \{answer\}\\
    \bottomrule
    \end{tabular}
    \caption{Prompts for generating and verifying responses. Qwen generates high-quality direct responses, while QwQ generates long-form reasoning responses and performs verification.} 
    \label{tab:ResGen}
\end{table*}

\clearpage

\section{Algorithm for comparing instruction scores}
\label{sec:appendixb}
\begin{algorithm}[h]
\caption{Algorithm for comparing instruction scores}
\begin{algorithmic}[1]
\State \textbf{Input}: item1: the score set of the first instruction; item2: the score set of the second instruction (The score set consists of three evaluation metrics—quality, complexity, and RelevanceToMedicine—as well as a boolean indicator MentionSpecificDetails denoting whether the response includes specific details.)
\State \textbf{Output}: item1 or item2 or None

\If{item1.MentionSpecificDetails \textbf{XOR} item2.MentionSpecificDetails}
    \If{item1.MentionSpecificDetails}
        \State \Return item2
    \Else
        \State \Return item1
    \EndIf
\EndIf

\If{item1.MentionSpecificDetails \textbf{AND} item2.MentionSpecificDetails}
    \State \Return None
\EndIf

\State score1 $\gets$ item1.quality + item1.complexity + item1.RelevanceToMedicine
\State score2 $\gets$ item2.quality + item2.complexity + item2.RelevanceToMedicine
\If{score1 $\neq$ score2}
    \State \Return item1 \textbf{ if } score1 $>$ score2 \textbf{ else } item2
\EndIf

\State score1 $\gets$ item1.quality + item1.complexity
\State score2 $\gets$ item2.quality + item2.complexity
\If{score1 $\neq$ score2}
    \State \Return item1 \textbf{ if } score1 $>$ score2 \textbf{ else } item2
\EndIf

\If{item1.quality $\neq$ item2.quality}
    \State \Return item1 \textbf{ if } item1.quality $>$ item2.quality \textbf{ else } item2
\EndIf

\State \Return randomChoice(\{item1, item2\})

\end{algorithmic}
\label{alg:compare_instruction_score}
\end{algorithm}

\section{Medical Knowledge Tree Classification Diagram}
\label{sec:appendixc}

\begin{figure}[ht]
\centering
\begin{tikzpicture}[
  grow=right, 
  level 1/.style={level distance=45mm, sibling distance=10mm},
  edge from parent/.style={draw, bend left=30,-latex},
  every node/.style={draw,rectangle,align=center,rounded corners}
]

\node[fill=blue!5] {Medicine}
  child {node {Other Departments}}
  child {node {Otorhinolaryngology}}
  child {node {Pediatrics}}
  child {node {Obstetrics and Gynecology}}
  child {node {Surgery}}
  child {node {Internal Medicine}};
\end{tikzpicture}
\caption{Hierarchical tree structure of Medicine and its subsets.}
\label{fig:tree}
\end{figure}
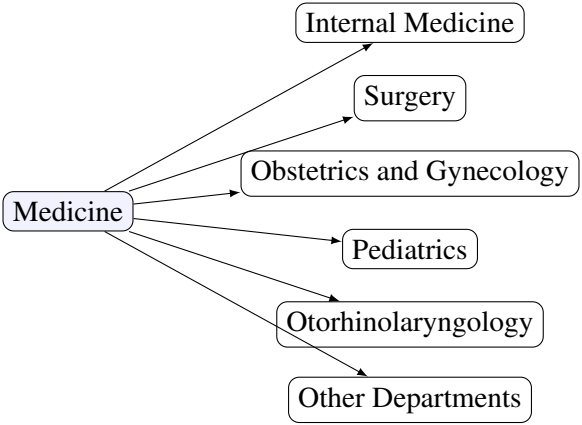

\begin{figure}[ht]
\centering
\begin{tikzpicture}[
  grow=right, 
  level 1/.style={level distance=45mm, sibling distance=10mm},
  edge from parent/.style={draw,bend left=30,-latex},
  every node/.style={draw,rectangle,align=center,rounded corners}
]
\node[fill=blue!5] {Internal Medicine}
    child {node {Neurology}}
    child {node {Rheumatology and Immunology}}
    child {node {Nephrology}}
    child {node {Hematology}}
    child {node {Gastroenterology}}
    child {node {Endocrinology}}
    child {node {Cardiology}}
    child {node {Respiratory and Critical Care Medicine}};

\end{tikzpicture}
\caption{Hierarchical tree structure of Internal Medicine and its subsets.}
\label{fig:tree}
\end{figure}
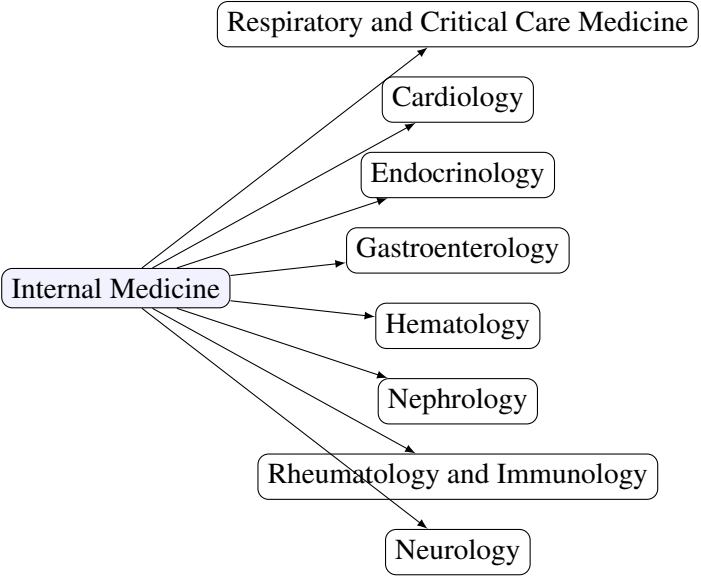

\begin{figure}[ht]
\centering
\begin{tikzpicture}[
  grow=right, 
  level 1/.style={level distance=45mm, sibling distance=10mm},
  edge from parent/.style={draw,bend left=30,-latex},
  every node/.style={draw,rectangle,align=center,rounded corners}
]
\node[fill=blue!5] {Surgery}
    child {node {Breast Surgery}}
    child {node {Thyroid Surgery}}
    child {node {Burns and Plastic Surgery}}
    child {node {Neurosurgery}}
    child {node {Orthopedic Surgery}}
    child {node {Thoracic Surgery}}
    child {node {Cardiovascular Surgery}}
    child {node {Urology}}
    child {node {Hepatobiliary Surgery}}
    child {node {Gastrointestinal Surgery}};

\end{tikzpicture}
\caption{Hierarchical tree structure of Surgery and its subsets.}
\label{fig:tree}
\end{figure}
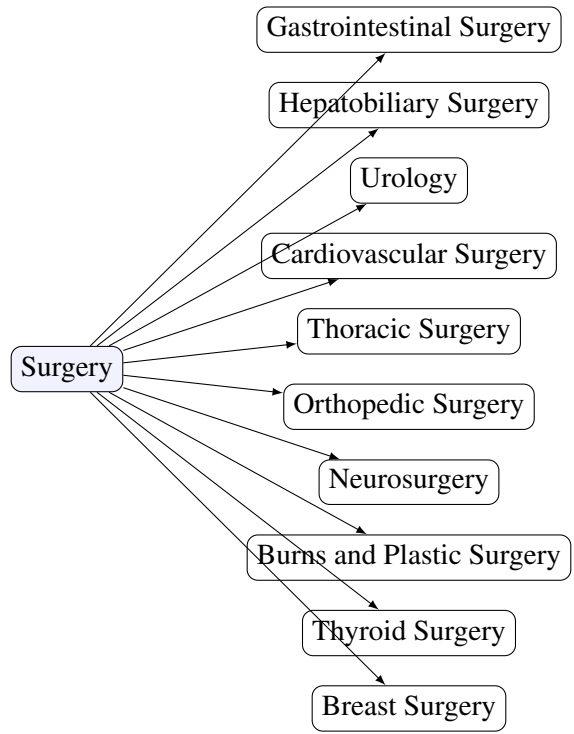

\begin{figure}[ht]
\centering
\begin{tikzpicture}[
  grow=right, 
  level 1/.style={level distance=45mm, sibling distance=10mm},
  edge from parent/.style={draw,bend left=30,-latex},
  every node/.style={draw,rectangle,align=center,rounded corners}
]
\node[fill=blue!5] {Obstetrics and Gynecology}
    child {node {Obstetrics}}
    child {node {Gynecology}};

\end{tikzpicture}
\caption{Hierarchical tree structure of Obstetrics and Gynecology and its subsets.}
\label{fig:tree}
\end{figure}
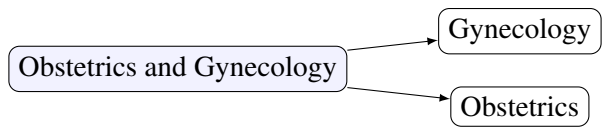

\begin{figure}[ht]
\centering
\begin{tikzpicture}[
  grow=right, 
  level 1/.style={level distance=45mm, sibling distance=10mm},
  edge from parent/.style={draw,bend left=30,-latex},
  every node/.style={draw,rectangle,align=center,rounded corners}
]
\node[fill=blue!5] {Pediatrics}
    child {node {Pediatric Surgery}}
    child {node {Pediatric Internal Medicine}};

\end{tikzpicture}
\caption{Hierarchical tree structure of Pediatrics and its subsets.}
\label{fig:tree}
\end{figure}
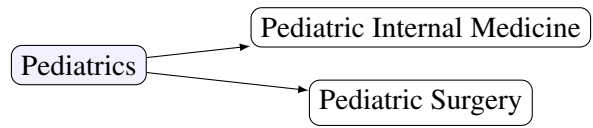

\begin{figure}[ht]
\centering
\begin{tikzpicture}[
  grow=right, 
  level 1/.style={level distance=45mm, sibling distance=10mm},
  edge from parent/.style={draw,bend left=30,-latex},
  every node/.style={draw,rectangle,align=center,rounded corners}
]
\node[fill=blue!5] {Otorhinolaryngology}
    child {node {Otorhinolaryngology}}
    child {node {Ophthalmology}}
    child {node {Ophthalmology}};

\end{tikzpicture}
\caption{Hierarchical tree structure of Otorhinolaryngology and its subsets.}
\label{fig:tree}
\end{figure}
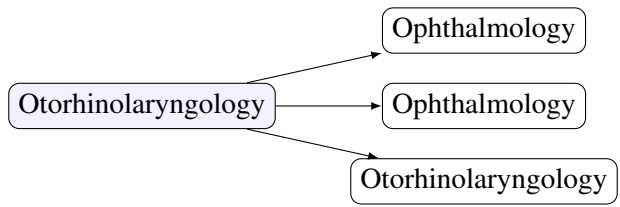

\begin{figure}[ht]
\centering
\begin{tikzpicture}[
  grow=right, 
  level 1/.style={level distance=45mm, sibling distance=10mm},
  edge from parent/.style={draw,bend left=30,-latex},
  every node/.style={draw,rectangle,align=center,rounded corners}
]
\node[fill=blue!5] {Other Departments}
    child {node {Traditional Chinese Medicine (TCM)}}
    child {node {Anesthesiology}}
    child {node {Rehabilitation Medicine}}
    child {node {Dermatology and Venereology}};

\end{tikzpicture}
\caption{Hierarchical tree structure of Other Departments and its subsets.}
\label{fig:tree}
\end{figure}
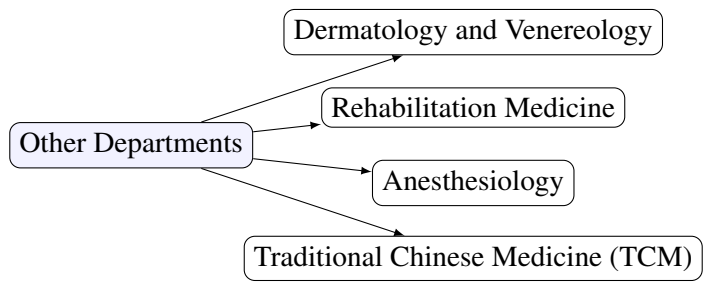

\clearpage

\section{Medical Data Classification Prompts}
\label{sec:appendixd}
\begin{table*}[ht]

\renewcommand{\arraystretch}{1.1}

\small
\centering
    \begin{tabular}{p{1.0\textwidth}}
    \toprule
    \textbf{\textsc{Medical Data Classification}}\\
    \midrule
    You are a professional doctor who can classify a dialogue. Please read the following dialogue and determine which hospital department it belongs to based on its main content. The available department categories and their brief descriptions are as follows:\\
    Internal Medicine: Diseases related to internal organs, such as heart disease, stomach problems, diabetes, etc.\\
    Surgery: Diseases treated with surgical interventions, such as fractures, tumor removal, etc.\\
    Obstetrics and Gynecology: Diseases and health issues related to the female reproductive system, pregnancy, and childbirth.\\
    Pediatrics: Health and disease issues related to children.\\
    Otorhinolaryngology (ENT): Diseases related to the ears, nose, throat, eyes, and oral cavity.\\
    Other Departments: Diseases or issues that do not fall under any of the above departments.\\
    Please choose the most appropriate department from the list above and output only the department name, without any additional explanation.\\
    Here is the dialogue: \\
    Patient: \{instruction\} \\
    Doctor: \{response\}\\
    Output: \\
    \bottomrule
    \end{tabular}
    \caption{Prompt for further segmentation of medical data.} 
    \label{tab:MedCls}
\end{table*}

\begin{table*}[ht]

\renewcommand{\arraystretch}{1.1}

\small
\centering
    \begin{tabular}{p{1.0\textwidth}}
    \toprule
    \textbf{\textsc{Internal Medicine Data Classification}}\\
    \midrule
    You are a professional doctor who can classify a dialogue. Please read the following dialogue and determine which specific department within Internal Medicine it belongs to based on its content. The available department categories and their brief descriptions are as follows:\\
    Respiratory and Critical Care Medicine: Treats respiratory system diseases such as asthma, pneumonia, chronic obstructive pulmonary disease (COPD), and provides care for critically ill patients.\\
    Cardiology: Covers diseases related to the heart and blood vessels, such as hypertension, coronary artery disease, arrhythmias, etc.\\
    Endocrinology: Focuses on diseases related to endocrine glands, such as diabetes, thyroid disorders, metabolic disorders, etc.\\
    Gastroenterology: Involves diseases of the digestive system, such as gastric ulcers, hepatitis, enteritis, etc.\\
    Hematology: Addresses blood-related diseases, such as anemia, leukemia, lymphoma, etc.\\
    Nephrology: Involves kidney diseases, such as nephritis, renal failure, uremia, etc.\\
    Rheumatology and Immunology: Deals with rheumatic diseases and immune system disorders, such as rheumatoid arthritis, systemic lupus erythematous, etc.\\
    Neurology: Focuses on diseases of the nervous system, such as stroke, epilepsy, Parkinson's disease, etc.\\
    If the dialogue content does not pertain to any of the above departments, please output 'None'.\\
    Please choose the most appropriate department from the list above and output only the department name, without any additional explanation.\\
    Here is the dialogue: \\
    Patient: \{instruction\} \\
    Doctor: \{response\}\\
    Output: \\
    \bottomrule
    \end{tabular}
    \caption{Prompt for further segmentation of internal medicine data.} 
    \label{tab:InterCls}
\end{table*}
\begin{table*}[ht]

\renewcommand{\arraystretch}{1.1}

\small
\centering
    \begin{tabular}{p{1.0\textwidth}}
    \toprule
    \textbf{\textsc{Surgical Data Classification}}\\
    \midrule
    You are a professional doctor who can classify a dialogue. Please read the following dialogue and determine which specific department within Surgery it belongs to based on its content. The available department categories and their brief descriptions are as follows:\\
    Gastrointestinal Surgery: Mainly involves surgical issues of the digestive system, such as stomach cancer, intestinal obstruction, etc.\\
    Hepatobiliary Surgery: Primarily involves surgical issues related to the liver, gallbladder, and pancreas, such as liver cancer, cholecystitis, pancreatic tumors, etc.\\
    Urology: Involves surgical issues of the urinary system, such as kidney stones, bladder tumors, benign prostatic hyperplasia, etc.\\
    Cardiovascular Surgery: Focuses on surgical diseases of the heart and major blood vessels, such as heart valve disease, aortic aneurysm, etc.\\
    Thoracic Surgery: Covers surgical procedures involving chest organs (lungs, esophagus, etc.), such as lung cancer and esophageal cancer.\\
    Orthopedic Surgery: Involves surgical diseases of bones, joints, and related structures, such as fractures, arthritis, herniated discs, etc.\\
    Neurosurgery: Handles surgical issues of the central and peripheral nervous systems, such as brain tumors, cerebral hemorrhage, spinal stenosis, etc.\\
    Burns and Plastic Surgery: Primarily responsible for burn treatment and plastic surgery procedures, such as burn scar repair and facial reconstruction, etc.\\
    Thyroid Surgery: Deals with surgical diseases of the thyroid and related glands, such as thyroid nodules and thyroid tumors, etc.\\
    Breast Surgery: Involves surgical treatment of breast diseases, such as breast hyperplasia and breast cancer.\\
    If the dialogue content does not pertain to any of the above departments, please output 'None'.\\
    Please choose the most appropriate department from the list above and output only the department name, without any additional explanation.\\
    Here is the dialogue: \\
    Patient: \{instruction\} \\
    Doctor: \{response\}\\
    Output: \\
    \bottomrule
    \end{tabular}
    \caption{Prompt for further segmentation of surgical data.} 
    \label{tab:MedCls}
\end{table*}
\begin{table*}[ht]

\renewcommand{\arraystretch}{1.1}

\small
\centering
    \begin{tabular}{p{1.0\textwidth}}
    \toprule
    \textbf{\textsc{Obstetrics and gynecology data classification}}\\
    \midrule
    You are a professional doctor who can classify a dialogue. Please read the following dialogue and determine which sub-department within Obstetrics and Gynecology it belongs to based on its content. The available sub-departments and their brief descriptions are as follows:\\
    Gynecology: Covers diseases and health issues related to the female reproductive system, such as menstrual disorders, infertility, uterine diseases, etc.\\
    Obstetrics: Involves pregnancy, childbirth, and postpartum care, such as pregnancy health management, delivery processes, and postpartum recovery.\\
    If the dialogue content does not pertain to any of the above sub-departments, please output 'None'.\\
    Please choose the most appropriate department from the list above and output only the department name, without any additional explanation.\\
    Here is the dialogue: \\
    Patient: \{instruction\} \\
    Doctor: \{response\}\\
    Output: \\
    \bottomrule
    \end{tabular}
    \caption{Prompt for further segmentation of obstetrics and gynecology data.} 
    \label{tab:MedCls}
\end{table*}
\begin{table*}[ht]

\renewcommand{\arraystretch}{1.1}

\small
\centering
    \begin{tabular}{p{1.0\textwidth}}
    \toprule
    \textbf{\textsc{Pediatric Data Classification}}\\
    \midrule
    You are a professional doctor who can classify a dialogue. Please read the following dialogue and determine which specific department within Pediatrics it belongs to based on its content. The available sub-departments and their brief descriptions are as follows:\\
    Pediatric Internal Medicine: Involves the diagnosis and treatment of internal diseases in children, including respiratory diseases, digestive system diseases, endocrine disorders, etc., such as colds, asthma, gastrointestinal issues, etc.\\
    Pediatric Surgery: Involves the diagnosis and treatment of surgical diseases in children, including trauma, congenital deformities, urinary system issues, etc., such as fractures, hernias, etc.\\
    If the dialogue content does not pertain to any of the above sub-departments, please output 'None'.\\
    Please choose the most appropriate department from the list above and output only the department name, without any additional explanation.\\
    Here is the dialogue: \\
    Patient: \{instruction\} \\
    Doctor: \{response\}\\
    Output: \\
    \bottomrule
    \end{tabular}
    \caption{Prompt for further segmentation of pediatric data.} 
    \label{tab:MedCls}
\end{table*}
\begin{table*}[ht]

\renewcommand{\arraystretch}{1.1}

\small
\centering
    \begin{tabular}{p{1.0\textwidth}}
    \toprule
    \textbf{\textsc{Otorhinolaryngology Data Classification}}\\
    \midrule
    You are a professional doctor who can classify a dialogue. Please read the following dialogue and determine which specific department within Otorhinolaryngology (ENT) it belongs to based on its content. The available sub-departments and their brief descriptions are as follows:\\
    Otorhinolaryngology (ENT): Involves diseases and issues related to the ears, nose, and throat, such as tinnitus, nasal congestion, sore throat, etc.\\
    Ophthalmology: Focuses on eye health issues, such as blurred vision, eye pain, dry eye, etc.\\
    Dentistry (Oral Medicine): Deals with issues related to the mouth, teeth, gums, etc., such as toothaches, mouth ulcers, gum bleeding, etc.\\
    If the dialogue content does not pertain to any of the above sub-departments, please output 'None'.\\
    Please choose the most appropriate department from the list above and output only the department name, without any additional explanation.\\
    Here is the dialogue: \\
    Patient: \{instruction\} \\
    Doctor: \{response\}\\
    Output: \\
    \bottomrule
    \end{tabular}
    \caption{Prompt for further division of ENT data.} 
    \label{tab:MedCls}
\end{table*}
\begin{table*}[ht]

\renewcommand{\arraystretch}{1.1}

\small
\centering
    \begin{tabular}{p{1.0\textwidth}}
    \toprule
    \textbf{\textsc{Other Department Data Classification}}\\
    \midrule
    You are a professional doctor who can classify a dialogue. Please read the following dialogue and determine which hospital department it belongs to based on its main content. The available department categories and their brief descriptions are as follows:\\
    Dermatology and Venereology: Primarily deals with skin diseases, sexually transmitted diseases, and related skin issues.\\
    Rehabilitation Medicine: Focuses on restoring patients' physical functions, treating sports injuries, and post-surgical rehabilitation.\\
    Anesthesiology: Mainly responsible for anesthesia management during surgeries, including general anesthesia and local anesthesia.\\
    Traditional Chinese Medicine (TCM): Uses traditional Chinese medicine theories and methods to treat diseases, including acupuncture, herbal medicine, and massage (Tui Na).\\
    If the dialogue content does not pertain to any of the above departments, please output 'None'.\\
    Please choose the most appropriate department from the list above and output only the department name, without any additional explanation.\\
    Here is the dialogue: \\
    Patient: \{instruction\} \\
    Doctor: \{response\}\\
    Output: \\
    \bottomrule
    \end{tabular}
    \caption{Prompt for further division of data of other departments.} 
    \label{tab:MedCls}
\end{table*}

\clearpage

\section{Statistics of the FineMed dataset}
\label{sec:appendixe}
\begin{table}[h]
  \centering
  \begin{tabular}{lcccc}
    \hline
    \textbf{Medical Specialty} & \textbf{Count} & \textbf{Quality} & \textbf{Complexity} &  \\
    \hline
    Respiratory and Critical Care Medicine          & 4766 & 8.40 & 6.42  \\
    Cardiology       & 10152 & 8.40 & 6.38  \\
    Endocrinology       & 11477 & 8.39 & 6.43 &  \\
    Gastroenterology & 7102 & 8.34 &  6.19 &   \\
    Hematology & 2068 & 8.47 &  6.51 &   \\
    Nephrology & 1728 & 8.40 &  6.46 &   \\
    Rheumatology and Immunology & 4662 & 8.40 &  6.42 &   \\
    Neurology & 3390 & 8.51 &  6.67 &   \\
    Gastrointestinal Surgery & 831 & 8.53 &  6.68 &   \\
    Hepatobiliary Surgery & 422 & 8.54 &  6.57 &   \\
    Urology & 431 & 8.53 &  6.78 &   \\
    Cardiovascular Surgery & 444 & 8.50 &  6.75 &   \\
    Thoracic Surgery & 349 & 8.58 &  6.81 &   \\
    Orthopedic Surgery & 4818 & 8.34 &  6.30 &   \\
    Neurosurgery & 1208 & 8.46 &  6.71 &   \\
    Burns and Plastic Surgery & 2589 & 8.18 &  6.14 &   \\
    Thyroid Surgery & 91 & 8.70 &  6.70 &   \\
    Breast Surgery & 1372 & 8.39 &  6.48 &   \\
    Gynecology & 8667 & 8.36 &  6.37 &   \\
    Obstetrics & 7559 & 8.36 &  6.24 &   \\
    Pediatric Internal Medicine & 1729 & 8.40 &  6.26 &   \\
    Pediatric Surgery & 298 & 8.49 &  6.59 &   \\
    Otorhinolaryngology  & 4923 & 8.27 &  6.06 &   \\
    Ophthalmology & 736 & 8.25 &  5.79 &   \\
    Dentistry (Oral Medicine) & 2128 & 8.28 &  5.70 &   \\
    Dermatology and Venereology & 10171 & 8.22 &  5.87 &   \\
    Rehabilitation Medicine & 7806 & 8.10 &  5.99    \\
    Anesthesiology & 514 & 8.24 &  6.36    \\
    Traditional Chinese Medicine (TCM) & 2869 & 8.00 &  6.26    \\

    \hdashline
    Internal Medicine & 78665 & 8.36 &  6.36    \\
    Surgery & 23541 & 8.32 &  6.37 &   \\
    Obstetrics and Gynecology & 21227 & 8.33 &  6.31 &   \\
    Pediatrics & 14253 & 8.26 &  6.26 &   \\
    Otorhinolaryngology & 8519 & 8.28 &  5.96 &   \\
    Other Departments & 158629 & 8.07 &  6.16 &   \\

    \hdashline
    Medical & 304858 & 8.20 &  6.24 &   \\
    DPO Dataset & 32919 & 8.89 &  8.03 &   \\
    \hline
  \end{tabular}
  \caption{
    Summary of key statistics and attributes of the FineMed dataset. \textbf{Quality} and \textbf{Complexity} refer to the average quality score and average complexity score, respectively. The table is divided into three sections: the first presents data from third-level departments, the second from second-level departments, and the third from first-level departments. Note that data with complexity score of 8 or higher have been extracted into the DPO dataset, so the average complexity score of the remaining data are comparatively lower.
  }
  \label{tab:statistics}
\end{table}

\clearpage

\section{Case Study on the Application of DPO and TTT}
\label{sec:appendixf}

\begin{figure}[h]
\centering
  \includegraphics[width=\textwidth]{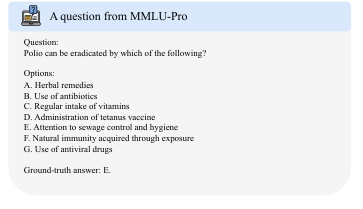}
  \label{fig:caseque}
\end{figure}

\begin{figure}[h]
\centering
  \includegraphics[width=\textwidth]{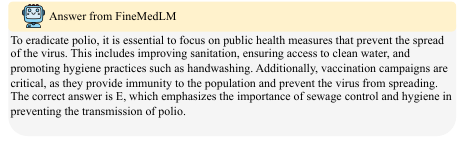}
  \label{fig:case1}
\end{figure}

\begin{figure}[h]
\centering
  \includegraphics[width=\textwidth]{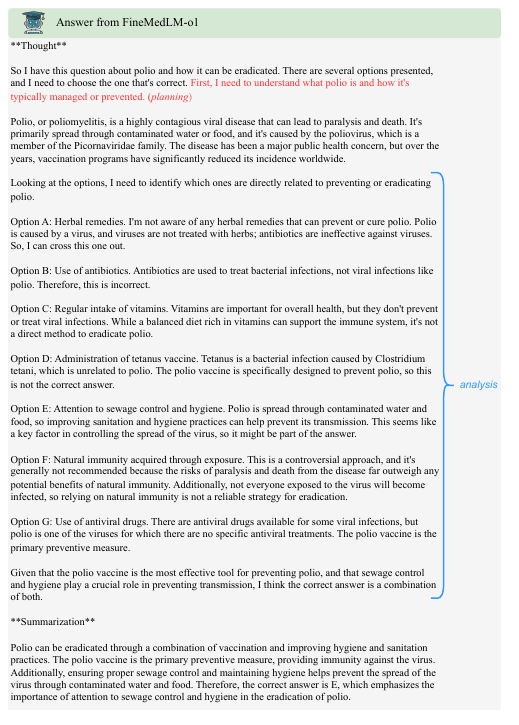}
  \label{fig:case2}
\end{figure}

\begin{figure}[h]
\centering
  \includegraphics[width=\textwidth]{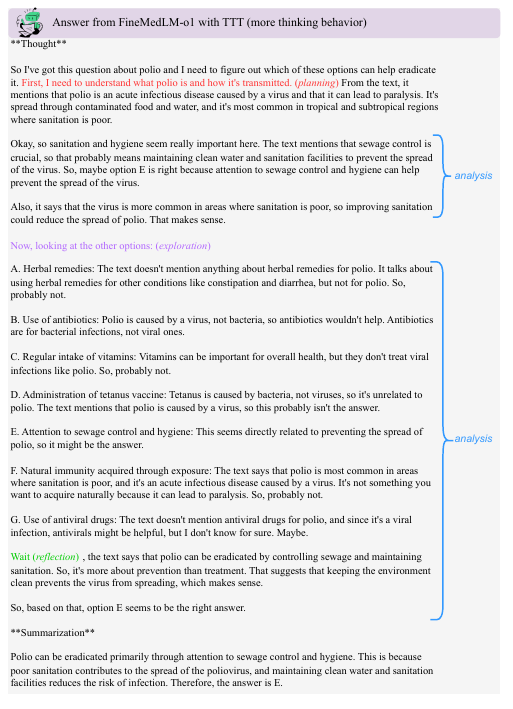}
  \label{fig:case3}
\end{figure}

\clearpage




\section{Comparative Results on Internal Medicine and Endocrinology Benchmarks}
\label{sec:appendixg}

\renewcommand{\arraystretch}{1.2} 
\begin{table}[ht]
  \centering
  \resizebox{\textwidth}{!}{%
    \begin{tabular}{lcccccc}
      \hline
      \textbf{Model}           & \textbf{Internal Medicine} & \textbf{Endocrinology} \\
      \hline
      HuatuoGPT2-7B\citep{chen2023huatuogpt} & 54.77 & 42.92  \\
      Medical-Llama3-8B  & 42.33 & 40.17  \\
      Llama3.1-8B\citep{dubey2024llama} & 40.77 & 38.99 \\
      FineMedLM-s1   & 55.01 & 43.25  \\
      FineMedLM-s2   &  60.55 &  46.33   \\
      FineMedLM   &  \textbf{62.50} &  \textbf{50.84}   \\
      \hline
    \end{tabular}
  }
  \caption{
    Main results on the internal medicine and endocrinology benchmarks. These benchmarks are constructed from private datasets collected within the hospital. FineMedLM-s1 and FineMedLM-s2 refer to models fine-tuned in the first and second stages, respectively.
  }
  \label{tab:privatebenchmark}
  \label{tab:privatebenchmark}
\end{table}



\clearpage

\end{document}